\documentclass[10pt,twocolumn,letterpaper]{article}

\usepackage[pagenumbers]{cvpr} % To force page numbers, e.g. for an arXiv version

\newcommand{\mysub}[1]{\textsubscript{#1}}
\usepackage{multirow}
\definecolor{cvprblue}{rgb}{0.21,0.49,0.74}
\definecolor{azure}{rgb}{0.0, 0.5, 1.0}
\usepackage[pagebackref,breaklinks,colorlinks,allcolors=cvprblue]{hyperref}

\def\paperID{1736} % *** Enter the Paper ID here
\def\confName{CVPR}
\def\confYear{2025}

\begin{document}
\title{LayoutDiT: Exploring Content-Graphic Balance in Layout Generation with Diffusion Transformer}

\author{
    Yu Li \textsuperscript{1}\footnotemark[1] \quad 
    Yifan Chen \textsuperscript{1}\footnotemark[1] \quad 
    Gongye Liu \textsuperscript{1} \quad 
    Fei Yin \textsuperscript{1} \quad 
    Qingyan Bai \textsuperscript{3} \quad \\
    Jie Wu \textsuperscript{1} \quad
    Hongfa Wang \textsuperscript{2} \quad 
    Ruihang Chu \textsuperscript{1}\footnotemark[2] \quad
    Yujiu Yang \textsuperscript{1}\footnotemark[2] \quad \\
$^{1}$Tsinghua University \quad 
${^2}$Tencent \quad
${^3}$HKUST \quad\\
    {\tt\small \{li-yu24, lgy22, wujie24, ruihangchu \}@mails.tsinghua.edu.cn \quad chenyf20@tsinghua.org.cn}
    \\ 
    {\tt\small \
    feii.yin@foxmail.com \quad qbaiac@connect.ust.hk}
    \\ 
    {\tt\small 
    \
    hongfawang@tencent.com \quad yang.yujiu@sz.tsinghua.edu.cn}
}

\twocolumn [ %
{\renewcommand\twocolumn[1][]{#1}%
\maketitle
\begin{center}
    \centering
\includegraphics[width=1\linewidth]{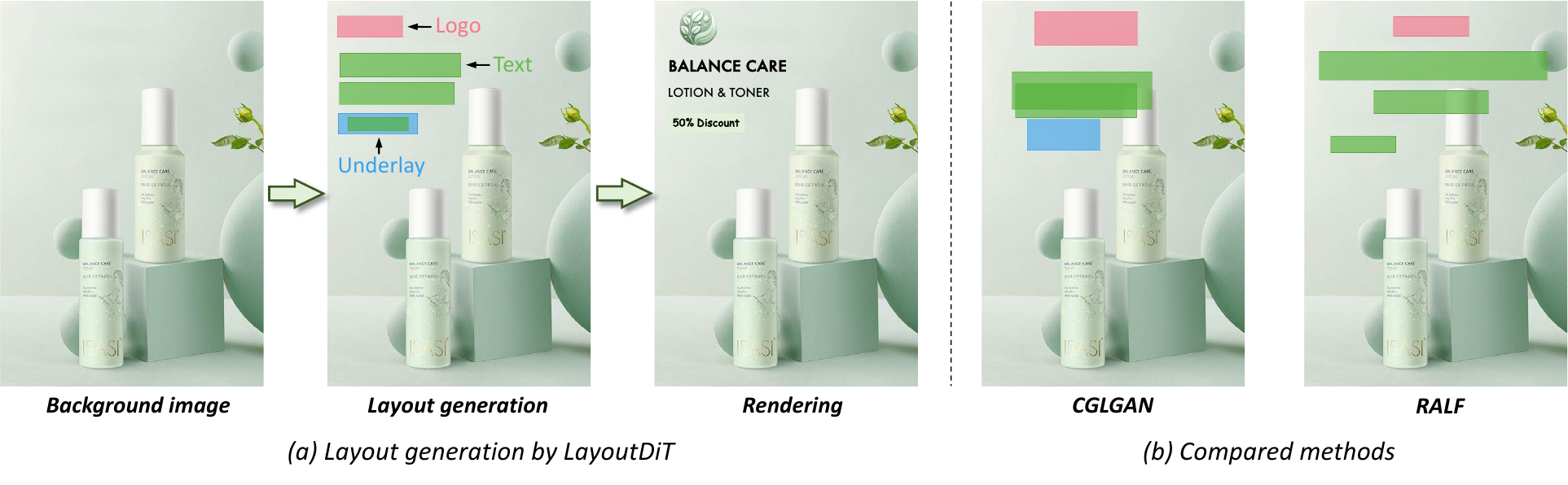}
\captionof{figure}{(a) Given a background image, our method generates reasonable and visually appealing layouts, which can be turned into a beautiful brand logo and advertising text via rendering.
(b) Layouts generated by SOTA methods suffer from issues such as blocking key image areas and overlapping with each other. In contrast, our approach produces a layout that shows well-structured graphic design and seamlessly integrates with the image content.
}
\label{fig:teaser}
\end{center}}
]

\renewcommand{\thefootnote}{\fnsymbol{footnote}}
\footnotetext[1]{Contribute equally.}
\footnotetext[2]{Corresponding author.}

\begin{abstract}
Layout generation is a foundation task of graphic design, which requires the integration of visual aesthetics and harmonious expression of content delivery. 
However, existing methods still face challenges in generating precise and visually appealing layouts, including blocking, overlapping, small-sized, or spatial misalignment.
We found that these methods overlook the crucial balance between learning content-aware and graphic-aware features. 
This oversight results in their limited ability to model the graphic structure of layouts and generate reasonable layout arrangements.
To address these challenges, we introduce LayoutDiT, an effective framework that balances content and graphic features to generate high-quality, visually appealing layouts.
Specifically, we first design an adaptive factor that optimizes the model's awareness of the layout generation space, balancing the model's performance in both content and graphic aspects.
Secondly, we introduce a graphic condition, the saliency bounding box, to bridge the modality difference between images in the visual domain and layouts in the geometric parameter domain.
In addition, we adapt a diffusion transformer model as the backbone, whose powerful generative capability ensures the quality of layout generation.
Benefiting from the properties of diffusion models, our method excels in constrained settings without introducing additional constraint modules.
Extensive experimental results demonstrate that our method achieves superior performance in both constrained and unconstrained settings, significantly outperforming existing methods. Our code is publicly available at \href{https://yuli0103.github.io/LayoutDiT.github.io/}{https://yuli0103.github.io/LayoutDiT.github.io/}.

\end{abstract}    
\section{Introduction}
\label{sec:intro}

Layout generation aims to generate an appropriate arrangement of elements, such as logos, text, and underlays, where each element is defined by a tuple of attributes like category, position, and size.
Logos and text serve as content carriers, conveying key information to the viewer.
Underlay elements are decorations below non-underlay elements (\eg, text), which cover the cluttered background to enhance readability.
Due to its versatility, this technique is widely applied in posters~\cite{guo2021vinci,o2015designscape}, documents~\cite{li2019layoutgan,patil2020read,zhong2019publaynet}, magazines~\cite{lee2020neural,yang2016automatic}, and user interfaces (UIs)~\cite{deka2017rico,hui2023unifying}.
Layout generation can be categorized into two types: content-agnostic and content-aware. 
Content-agnostic task generates layouts without considering the background image or its content, while content-aware layout generation aims to generate a fitting layout that harmonizes with the given background image. 
This paper focuses on the content-aware setting, where we generate high-quality layouts that not only follow a graphic structure consistent with human aesthetics but also seamlessly integrate with the image content.

For context-aware layout generation, generative models~\cite{goodfellow2020generative,ho2020denoising,rombach2022high,xia2022gan} have made notable progress, including generative adversarial networks (GANs)~\cite{hsu2023posterlayout,xu2023unsupervised,zheng2019content,zhou2022composition}, variational autoencoders (VAE)~\cite{cao2022geometry}, and transformer-based models~\cite{horita2023retrieval}.
However, we find that these models still struggle to accurately learn layout distributions, particularly in balancing well-structured graphic arrangements with content harmony. As showcased in Fig.~\ref{fig:teaser}(b), this learning imbalance often leads to critical issues such as blocking, overlapping, small-sized, and spatial misalignment.
To tackle these challenges, Diffusion Transformer (DiT) models~\cite{videoworldsimulators2024,chen2023pixart,esser2024scaling} have emerged as a promising solution. Unlike GAN-based and VAE-based methods which suffer from training instability and difficulty in modeling complex layout graphic structure, DiT framework has shown superior capabilities in sequence generation tasks~\cite{chen2024towards, levi2023dlt, inoue2023layoutdm}. Yet, existing diffusion transformers have focused on content-agnostic tasks,  lacking effective mechanisms to integrate content considerations for generating coherent layouts.

% current approaches \textcolor{blue}{fail to effectively balance the emphasis} on content-aware features of images and graphic features of layouts.
% Consequently, generating high-quality layouts remains a significant challenge.

Nevertheless, directly applying DiT architecture to content-aware layout generation still faces the challenge of balancing content-aware and graphic features.
We hypothesize that this limitation stems primarily from the lack of awareness for the layout generation space.
In content-agnostic tasks, the generation space naturally encompasses the entire canvas, since the model does not need to consider aligning with the background image.
While in content-aware tasks, the presence of a given image imposes constraints that compress the generation space. 
The model's limited perception of a compressed generation space leads to the generated layout elements with incorrect spatial positions and dimensions, finally showing imbalanced performance between content and graphic aspects.
To address this issue, we introduce a novel content-graphic balance factor to adaptively optimize the model's awareness of the generation space for different samples. This factor is predicted by a trainable module, serving as a dynamic modulator between layout representations and image features. Compared with the fixed factor, whose results tend to be biased towards either content or graphic aspects, our learnable paradigm improves coherence and layout quality by achieving optimal content-graphic balance.

Additionally, due to the modality difference between images in the visual domain and layouts in the geometric parameter domain, the model struggles to align geometric features effectively. 
Therefore, we propose saliency bounding boxes as an enhanced condition, which are extracted from the saliency map. 
These bounding boxes effectively extract geometric information from images and transform it into the same modality as layout elements, thus providing a more precise basis for geometric feature alignment.

By seamlessly incorporating these critical designs into an advanced Diffusion Transformer architecture, our integrated framework, termed as LayoutDiT, enables end-to-end high-quality layout generation. We evaluate our method on two public benchmarks PKU~\cite{hsu2023posterlayout} and CGL~\cite{zhou2022composition}.
Notably, our approach significant improves graphic metrics and content harmony, effectively mitigating issues of underlay misalignment and text overlap.
In addition, by leveraging diffusion models' characteristics, our framework handles constrained settings
effectively without introducing additional constraint modules.
Extensive experimental results show that our method excels in both unconstrained and constrained settings and consistently outperforms state-of-the-art methods.
Our main contributions are as follows:

\begin{itemize}[leftmargin=1em]
    \setlength{\itemsep}{0pt}

    \item We successfully apply a Diffusion Transformer model in content-aware layout generation to unleash its sequence modeling capability, which is the first of this kind to our best knowledge.

    \item We introduce a novel content-graphic balance factor to enhance model's awareness of layout generation space, as well as leveraging saliency bounding boxes to bridge the modality gap between images and layouts, finally improving performance in both content and graphic aspects.

    \item Extensive experiments, quantitatively and qualitatively, show that our method achieves SOTA performance under both unconstrained and constrained settings. 

\end{itemize}

\section{Related Work}

\subsection{Diffusion-based Content-agnostic Layout Generation}
Numerous studies on content-agnostic layout generation~\cite{arroyo2021variational,gupta2021layouttransformer,jiang2023layoutformer++,tang2023layoutnuwa} have incorporated diffusion models into this field.
PLay~\cite{cheng2023play} utilizes a conditional latent diffusion model to generate parametrically conditioned layouts in the vector graphic space. 
LayoutDM~\cite{inoue2023layoutdm} and LayoutDiffusion~\cite{zhang2023layoutdiffusion} are based on discrete diffusion models~\cite{austin2021structured}, treating both geometric parameters and category information as discrete data. 
Another study~\cite{chai2023layoutdm} employs DDPM to handle geometric parameters in continuous spaces while introducing category information as a condition. 
In contrast, DLT~\cite{levi2023dlt}  proposes training discrete and continuous data jointly through a diffusion layout transformer. 
LACE~\cite{chen2024towards} introduces a unified model that generates in a continuous feature space, enabling the integration of differentiable aesthetic constraint functions during training to guide the optimization of layout generation. 
Our research aims to expand the generative scope of diffusion models, focusing on exploring how to apply diffusion models to content-aware layout generation.

\begin{figure*}[t]
    \centering 
    \includegraphics[width=0.9\linewidth]{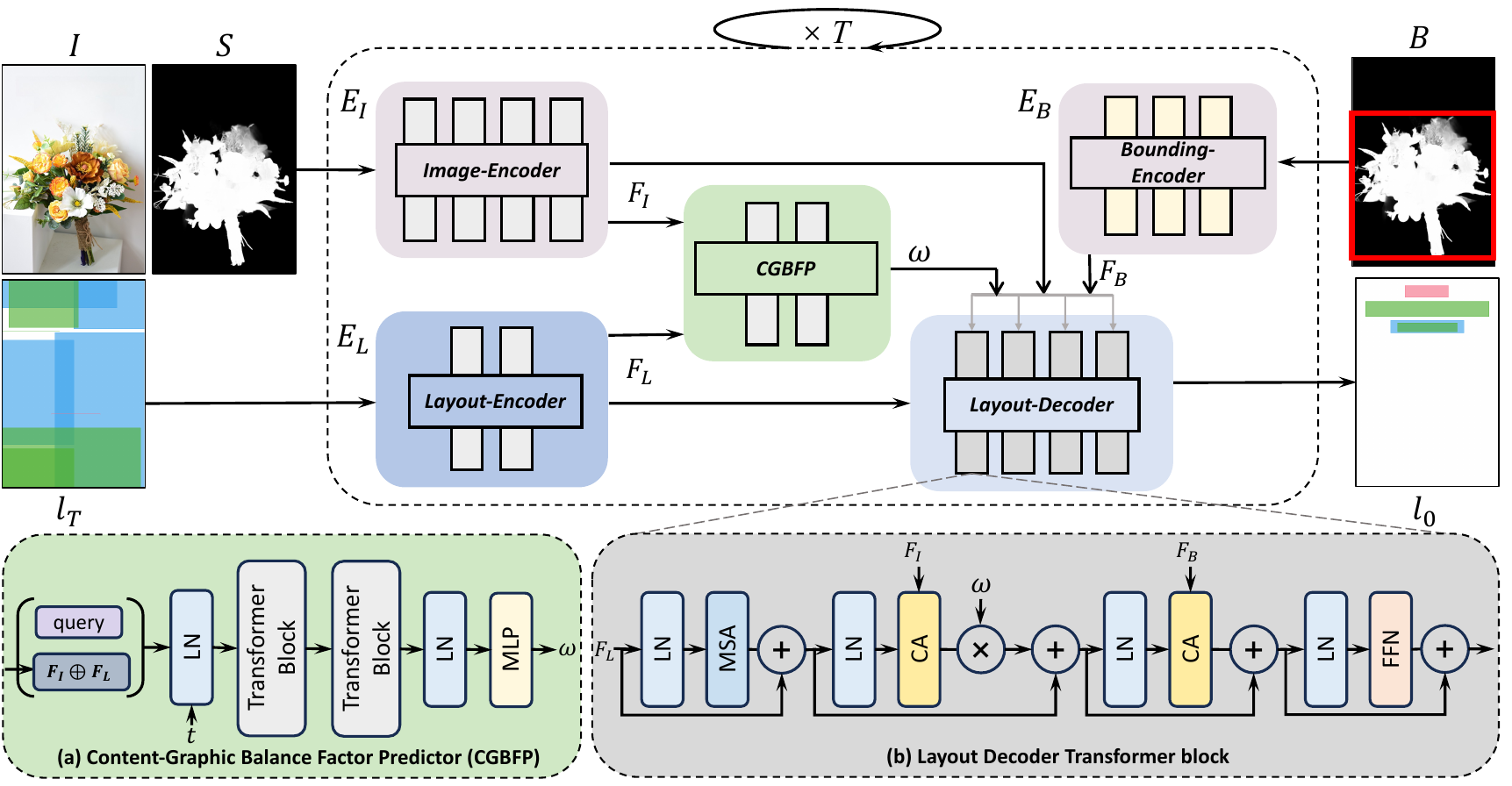}
    \caption{%
        The overview of our framework.
        The inputs are Gaussian noise $l_T$, image $I$ with its 
        saliency map $S$, and saliency bounding box $B$. The layout encoder and decoder serve as the main backbone, both composed of a series of transformer blocks.
        Image features $F_I$ and box 
        features $F_B$ are extracted by the image encoder and bounding encoder respectively, 
        and are incorporated into the backbone through cross-attention modules. The CGBFP 
        module takes $F_I$ and layout representations $F_L$ as inputs to predict a balance 
        factor $\omega$, which modulates the cross-attention interactions. Finally, the 
        framework generates high-quality and visually appealing layouts $l_0$.%
    }
    \label{fig:architecture}
    \vspace{-6pt}
\end{figure*}

\subsection{Content-aware Layout Generation}

Content-GAN~\cite{zheng2019content} was the first layout generation model to include both visual and textual content. 
Subsequently, CGL-GAN~\cite{zhou2022composition} focused on addressing the domain misalignment issue, pioneering subsequent research by introducing a saliency map.
ICVT~\cite{cao2022geometry} studied layout elements across visual and geometric parameter domains, using a geometry alignment module to align image features with layout representations.
The DS-GAN~\cite{hsu2023posterlayout} team explored the layout generation task using a distinct model architecture CNN-LSTM. 
To our best knowledge, RADM~\cite{li2023relation} was the first to apply the diffusion architecture to this task. However, unlike RADM, our approach utilizes the transformer model~\cite{vaswani2017attention} to model layout distributions.
RALF~\cite{horita2023retrieval} approached from a data perspective, identifying that misaligned underlay embellishment and text elements commonly found in prior studies were due to a lack of highly structured layout training data, hence proposing a retrieval-augmented approach to mitigate data scarcity issues.

While prior approaches have contributed significantly to layout generation, they face an inherent balance dilemma between learning content-aware and graphic features when handling diverse background images, resulting in various issues. In contrast, our LayoutDiT effectively addresses these issues, paving the way for generating high-quality layouts.

\section{Method}

Fig.~\ref{fig:architecture} illustrates the overview of our framework. We formulate layout generation as a denoising process and adopt a diffusion transformer architecture to capture layout patterns. Our method enables the model to effectively balance content perception and graphic structure modeling.

\subsection{Preliminary: Diffusion Model}\label{sec:3.1}

The forward diffusion process starts by sampling $x_{0}$ from the true data distribution, and then gradually adds Gaussian noise at each timestep $t$ to progressively corrupt the data:
\begin{equation}\label{eq:1}
\begin{split}
    q(x_t|x_{t-1})   &= \mathcal{N}(x_t; \sqrt{1 - \beta_{t}}x_{t-1}, \beta_{t}\mathrm{I}), \\
    q(x_{1:T}|x_{0}) &= \prod_{t=1}^{T}q(x_{t}|x_{t-1}),
\end{split}
\end{equation}
where $[{\beta_t}]_{t=0}^T$ denotes a constant variance schedule that controls the Gaussian noise at each step. We can obtain 
\begin{equation}
x_t = \sqrt{\overline{\alpha}_t}x_0 + \sqrt{1 - \overline{\alpha}_t}\epsilon,
\end{equation}
where $\alpha_{t} = 1 - \beta_{t}$, $\overline{\alpha_{t}} = \prod_{s=1}^{t}\alpha_{s}$, and $\epsilon \sim \mathcal{N}(0, \mathrm{I})$ is the random noise added at time-step $t$. The reverse process model is trained to invert the forward process by predicting the statistics of $p_{\theta}$ as :
\begin{equation}
p_{\theta}(x_{t-1}|x_{t}) = \mathcal{N}(x_{t-1}; \mu_\theta(x_{t}, t), \Sigma_\theta(x_{t}, t)).
\end{equation}
By setting $\Sigma_\theta(x_{t}, t) = \sigma_t^2 I$ and reparameterizing $\mu_\theta$ as a noise predictor network $\epsilon_{\theta}$~\cite{ho2020denoising}, the model is trained by minimizing the L2 loss between the predicted noise and the ground truth sampled Gaussian noise  $\epsilon$. The training object can be written as minimizing:
\begin{equation}
\min_{\theta} \mathbb{E}_{t,x_t,\epsilon} [ \lVert \epsilon - \epsilon_\theta(x_t, t) \rVert ^2 ].
\end{equation}

\subsection{Overview}\label{sec:3.2}

We formulate layout generation as a denoising process, which gradually transforms noise into coherent and precise layouts.
The layout is defined as a set of $N$ elements, denoted as $L = \{l_1, ..., l_N \} = \{(c_1, \mathrm{b}_1), ..., (c_N, \mathrm{b}_N) \}$. In this notation, $c_i \in \{0, 1, ..., C \}$ represents $C$ categories of layout elements, along with a special category denoted by $c_i = 0$ to signify an empty element. Additionally, $\mathrm{b}_i = (x_i, y_i, w_i, h_i) \in [0, 1]^4 $ indicates the normalized box.

The image conditions consist of a paired canvas $I \in \mathbb{R}^{H \times W \times 3}$ and its corresponding saliency map $S \in \mathbb{R}^{H \times W \times 1}$, where $H$ and $W$ denote the height and width, respectively. The saliency map $S$ is obtained from the canvas through an off-the-shelf detection method~\cite{qin2019basnet,qin2022highly}.

Fig.~\ref{fig:architecture} illustrates the overview of our framework. The layout encoder and decoder serve as the main backbone, both composed of a series of transformer blocks. While the encoder's blocks follow the standard architecture with self-attention mechanisms, the decoder's blocks are structured as shown in Fig.~\ref{fig:architecture} (b), incorporating both self-attention and cross-attention modules to interact with various conditions. 

The concatenated $I$ and $S$ serve as image conditions, which are processed by a ViT~\cite{dosovitskiy2020image}-based image encoder to obtain image features ($F_I$). Subsequently, $F_I$ and the layout representations $F_L$ from the layout encoder are fed into the CGBFP module to predict the balanced factor $\omega$, which modulates the interaction between layout representations and image features in the cross-attention module. Additionally, we introduce an enhanced condition called saliency bounding box (B), which is processed by a bounding encoder to obtain box features $F_B$. Similarly, $F_B$ serves as a condition for the second cross-attention module in the decoder blocks. The balanced factor and saliency bounding box will be detailed in Sec.~\ref{sec:3.3} and~\ref{sec:3.4}, respectively.

\subsection{Content-Graphic Balance}\label{sec:3.3}
We attribute a key reason for previous methods' issues (blocking, overlapping, small-sized, or misalignment) to their oversight of the layout generation space. In content-agnostic generation tasks~\cite{jiang2022coarse,kikuchi2021constrained,kong2022blt}, such unreasonable layout elements are typically avoided since the model does not need to consider aligning with the images. Therefore, the space of layout generation is the entire canvas. However, in content-aware tasks, the generation space is compressed to varying degrees across different samples due to the influence of image saliency regions. When the model is overly sensitive to the generation space, it tends to produce layouts with small-sized, overlapping, or misalignment issues in an attempt to find suitable positions within limited spaces, resulting in poor graphics performance. Conversely, insufficient awareness of the generation space leads to layouts that obstruct important regions in the images. We found that previous approaches suffer from improper perception of the layout generation space, which leads to imbalanced performance between content and graphics aspects.
\begin{figure*}[htbp]
    \centering
    \includegraphics[width=0.8\textwidth]{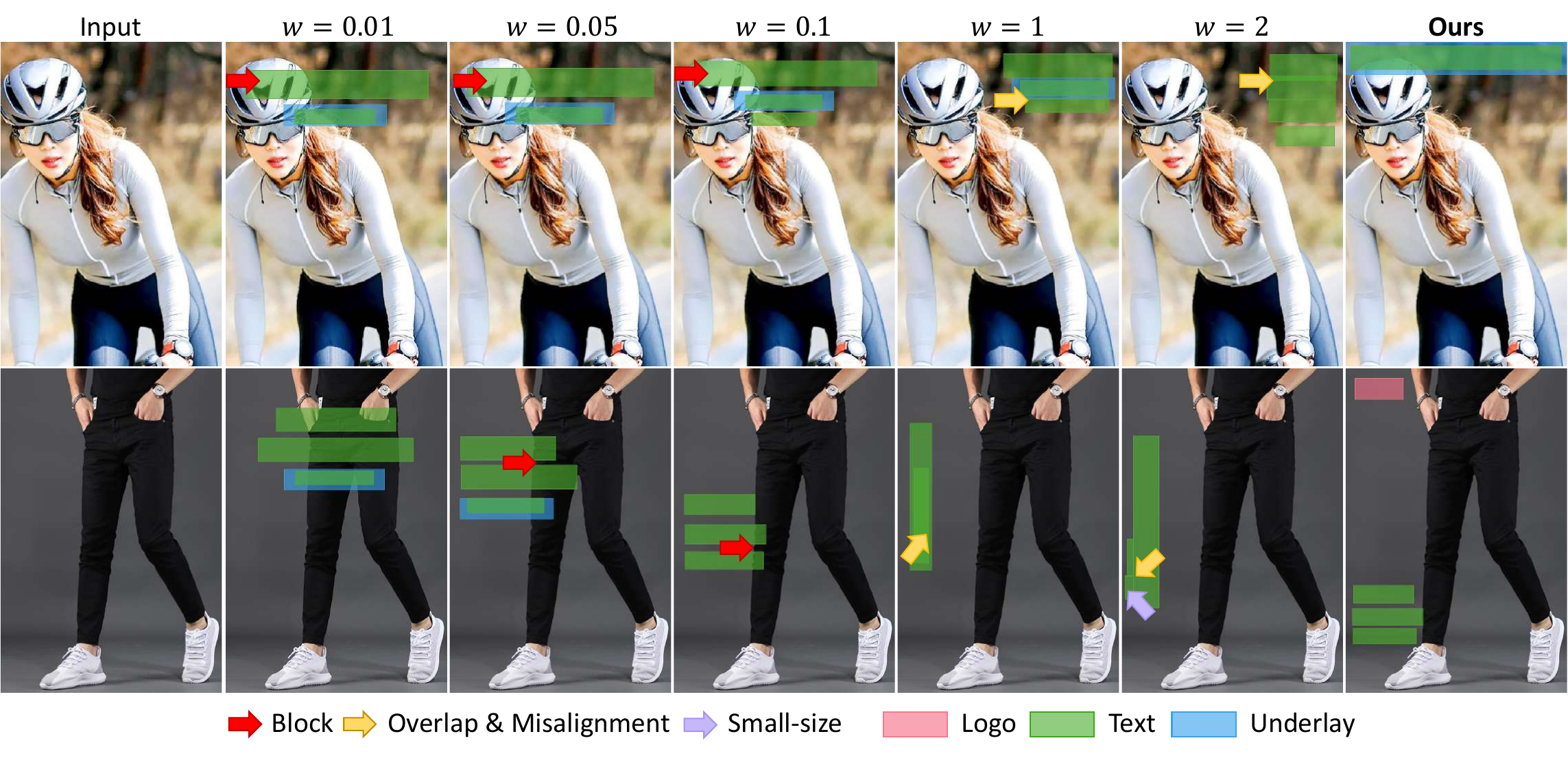}
    \caption{Exploration of the factor $w$ on PKU dataset. The factor $w$ enhances the model’s awareness of layout generation space, thereby achieving optimal content-graphic balance.
    In contrast, using a constant factor often leads to poor layout performance in either aspect.
    }
    \label{fig:pku_weight_uncon}
\end{figure*}
To address this, we introduce a content-graphic balance factor $\omega$ that optimizes the model's awareness of the layout generation space. Specifically, we employ a trainable CGBFP module (implemented by a query transformer network~\cite{li2023blip}), which takes image features $F_I$, layout representations $F_L$, and timestep $t$ as input to predict $\omega$:

\begin{equation}\label{eq: factor}
\omega = CGBFP\left(F_{I}, F_{L}, t\right)
\end{equation}

The factor $\omega$ is subsequently mapped to the intermediate layers of the transformer model through cross-attention, modulating the interaction between layout representations and image features:

\begin{equation}
    X^{''} = X^{'} + ~ \omega ~ * ~ \text{CA}(X^{'}, F_{I}, F_{I})
\end{equation}

Here, $X^{'}$ denotes the intermediate feature after the self-attention module, CA denotes cross-attention.

To demonstrate the advantages of our adaptive factor, we compared it with constant values (Fig.~\ref{fig:pku_weight_uncon}), with detailed quantitative results provided in Appendix. Results show that using constant $\omega$ leads to inappropriate layout arrangements, causing bias towards either content or graphics aspects. In contrast, our adaptive factor enables an effective balance between graphic and content-aware metrics.

\subsection{Saliency Bounding Box}\label{sec:3.4}

We found that the model's alignment of geometric features between the layout representations and image embeddings is insufficient, likely due to the difference in modalities, with images in the visual domain and layouts in the geometric parameter domain. Furthermore, we noted that the saliency map effectively retains the key shapes of the canvas while removing other irrelevant high-frequency details. 
Consequently, we employ a contour detection algorithm to process the saliency map, obtaining the saliency bounding box. Specifically, we detect the boundaries of salient areas in the saliency map by thresholding pixel values, where the saliency bounding area value is expressed as:
\begin{equation}
 B_{ij}=\left\{
\begin{array}{rcl}
0,       &      & {B_{ij} \leq s},\\
1,       &      & {B_{ij} > s},
\end{array} \right. 
\end{equation}
where $s$ is a threshold value and $i,j \geq 1$.
Based on this binary map, we then determine the bounding box using the minimum rectangle that encloses all regions where $B_{ij}=1$.

The bounding box is then encoded into $F_B$ through a bounding encoder implemented with a multi-layer perceptron (MLP). This step explicitly extracts geometric information from salient regions in images and transforms it into the same modality as layout elements, thus providing a more precise basis for geometric feature alignment.
Similar to Sec.~\ref{sec:3.3}, we employ cross attention to inject saliency bounding box features:

\begin{equation}
    X^{'''} = X^{''} + ~ \text{CA}(X^{''}, F_{B}, F_{B})
\end{equation}

Since each saliency map contains only a few bounding boxes, this approach introduces negligible computational overhead while achieving better performance.

\section{Experiment}

% \subsection{Settings}
\subsection{Datasets}\label{sec:4.1}
% \noindent \textbf{Datasets.} 
\begin{figure*}[ht]
\begin{center}
\centerline{\includegraphics[width=1\linewidth]{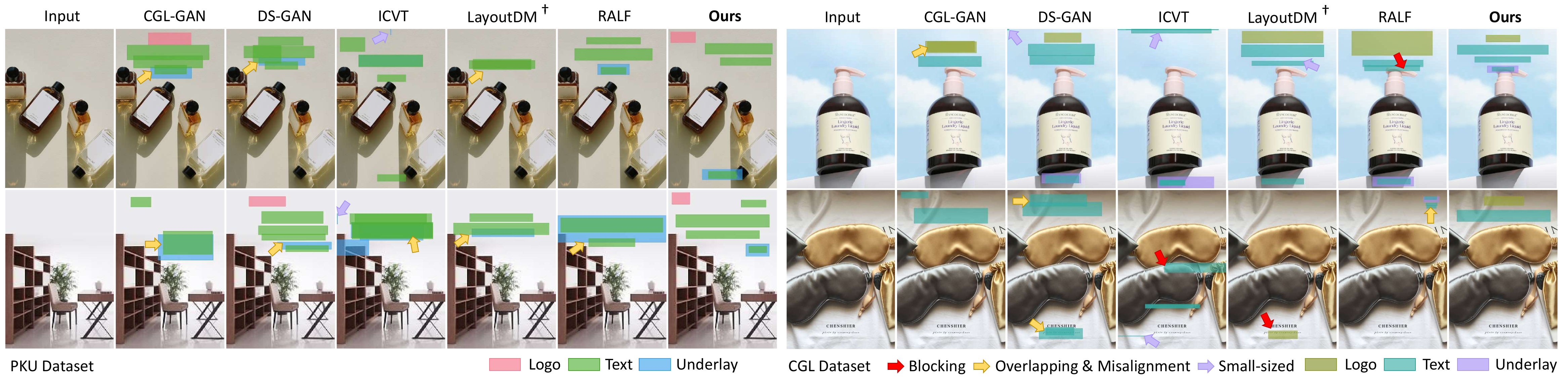}}
\vspace{-2pt}
\caption{
Qualitative comparison of unconstrained generation on PKU and CGL. Compared to other approaches, LayoutDiT more effectively handles issues of blocking, overlapping, misalignment and small-sized elements while preserving layout diversity.}
\label{fig:pku_cgl_un}
\end{center}
\vspace{-20pt}
\end{figure*}
We evaluated our method and other baselines on two publicly available e-commerce datasets: CGL~\cite{zhou2022composition} and PKU~\cite{hsu2023posterlayout}. Due to the absence of annotated test sets in the original datasets, we were unable to evaluate the constrained generation tasks, which require ground-truth layout attribute labels. Therefore, we followed RALF's~\cite{horita2023retrieval} setup by splitting the original training sets with a train/validation/test ratio of 8:1:1, and conducted experiments on both annotated and unannotated test sets. Due to the space limitations, we present detailed dataset analysis and processing procedures in Appendix, along with most results on the unannotated test sets.

\begin{table}[t]
	\centering
	\caption{Comparison of small-sized elements (area $<$ 0.1\% of canvas, or width/height $<$ 2\% of canvas) across different methods. }
    \renewcommand{\arraystretch}{1.0} 
    \resizebox{\linewidth}{!}{
        \begin{tabular}{ccccccc}
            \toprule
            & \multicolumn{6}{c}{Sma: Small-sized proportion (\%)} \\
            \cmidrule(lr){2-7}
            & DS-GAN & CGL-GAN & ICVT & LayoutDM$^{\dagger}$ & RALF & Ours \\
            \midrule
            CGL-unanno & 14.02 & 1.48 & 16.11 & 5.82 & 9.60 &\textbf{0.33}  \\
            CGL-anno& 13.00 & 1.56 & 9.02 & 2.09 & 2.03 &  \textbf{0.30} \\
            PKU-unanno & 0.21 & 0.26 & 19.96 & \textbf{0.00} & 0.17 & \textbf{0.00}  \\
            PKU-anno & 0.35 & 0.05 & 20.17 & \textbf{0.00} & 0.13 & \textbf{0.00} \\
            \bottomrule
        \end{tabular}
    }
	\label{tab:small-sized}
\end{table}
% Sma: Small-sized proportion

\subsection{Evaluation Metrics}
% We \textcolor{red}{evaluated} primarily from the following three dimensions.

\noindent \textbf{Graphic metrics.}
This part of the assessment focuses on measuring the graphic quality of the layout without considering the canvas. 
The $\text{Und}_\text{L}$ calculates the overlap ratio between each non-underlay element and an underlay element, then selects the maximum value among these ratios.
The $\text{Und}_{\text{S}}$ imposes stricter conditions, only including instances where the overlap ratio is 1.
The Overlap (\text{Ove}) represents the average IoU of all pairs of elements except for underlay elements. 
The Small-sized proportion (\text{Sma}) measures the ratio of small-sized elements in the generated layout.

\noindent \textbf{Content-aware metrics.}
This part primarily evaluates the harmony between the layout and the canvas. The Occlusion (\text{Occ}) calculates the average proportion of layout elements covering the saliency regions, identified through saliency maps. The Readability (\text{Rea}) assesses the non-flatness of regions containing plain text elements without underlay decoration by calculating the average gradient of pixels in the layout area within the image space. 

% 插入user study
\noindent \textbf{User study.}
We randomly selected 20 images to evaluate all methods. A group of volunteers assessed the eligibility of generated layouts and identified the best one per image. For each method, we computed the selection rate ($P_{sel}$) and preference rate ($P_{pref}$). The results are shown in Tab.~\ref{tab:user-study}.

\subsection{Implementation Details}

All experiments were conducted on an NVIDIA 4090 GPU. Our LayoutDiT was trained for 500 epochs with different configurations: on PKU (batch size 32, learning rate 1e-4) and on CGL (batch size 128, learning rate 2e-4). Based on RALF's replication setup, we retrained all benchmark methods (CGL-GAN, DS-GAN, ICVT, LayoutDM, and RALF) on our processed datasets for comparison. Given that RADM~\cite{li2023relation} incorporates textual information which is not available in all datasets, it presents challenges for fair comparison with other methods. Consequently, we excluded RADM from the result comparison. Specific network architecture details are provided in Appendix.

\begin{table}[t]
	\centering
	\caption{Results of user study. 
                }
    \renewcommand{\arraystretch}{1.0}
	\resizebox{\linewidth}{!}{
		\begin{tabular}{ccccccc}
			\toprule
			 & DS-GAN & CGL-GAN & ICVT & LayoutDM$^{\dagger}$ & RALF & Ours \\
			\midrule
			$P_{sel}$  &44.05 &30.00 &37.14 &24.29 &67.38 &\textbf{92.38} \\
			$P_{pref}$&7.38  &3.57  &2.14 &1.43 &18.81 & \textbf{66.67}  \\
			\bottomrule
		\end{tabular}
	}
	\label{tab:user-study}
\end{table}

\subsection{Unconstrained Generation}

\noindent \textbf{Qualitative comparison.} 
Fig.~\ref{fig:pku_cgl_un} presents the qualitative results on the PKU and CGL datasets, with different colored arrows highlighting specific issues.
While CGL-GAN and DS-GAN demonstrate general comprehension of image content, they exhibit significant issues with text overlap and underlay misalignment.
ICVT shows limitations in both content and graphic quality, potentially due to its VAE architecture's inherent training challenges.
LayoutDM struggles to accurately discern the geometric properties for different layout elements (e.g., improper text widths and logo positions).
RALF represents the previous SOTA model, benefiting significantly from retrieval augmentation to enhance results. Nevertheless, it still faces challenges in accurately modeling layout graphic structure. 
In contrast, our method successfully addresses these limitations observed in previous methods.

\begin{table*}[htbp]
\vspace{-6pt}
\centering
\caption{Comparison of unconstrained generation methods on PKU and CGL annotated datasets.}
\label{tab:un_all_annotated_quantitative}
\resizebox{\linewidth}{!}{
\begin{tabular}{lccccccccccc}
% \begin{tabularx}{\textwidth}{@{}Xccccccccccccccc@{}}
\toprule
\quad & \quad & \multicolumn{5}{c}{PKU annotated} & \multicolumn{5}{c}{CGL annotated} \\
\cmidrule(lr){3-7} \cmidrule(lr){8-12}
Method & Params   & \multicolumn{2}{c}{Content} & \multicolumn{3}{c}{Graphic} & \multicolumn{2}{c}{Content} & \multicolumn{3}{c}{Graphic} \\
\cmidrule(lr){3-4} \cmidrule(lr){5-7} \cmidrule(lr){8-9} \cmidrule(lr){10-12}

\quad & \quad  & Occ~$\downarrow$ & Rea~$\downarrow$  & Und\mysub{L}~$\uparrow$ & Und\mysub{S}~$\uparrow$ & Ove~$\downarrow$ & Occ~$\downarrow$ & Rea~$\downarrow$ & Und\mysub{L}~$\uparrow$ & Und\mysub{S}~$\uparrow$ & Ove~$\downarrow$ \\ 
\midrule
CGL-GAN\cite{zhou2022composition} &41M 
        & 0.176 & 0.0167 &0.745& 0.333  & 0.1071  
        & 0.196 & 0.0249 &0.729& 0.310  & 0.2574  \\

DS-GAN\cite{hsu2023posterlayout}  &30M    
        & 0.159 & 0.0161 & 0.828  & 0.575& 0.0353
        & 0.153 & 0.0198 & 0.886  & 0.545& 0.0415  \\

ICVT\cite{cao2022geometry}  &50M      
        & 0.280 & 0.0205 &0.446 &0.296  & 0.3155 
        & 0.227 & 0.0267 &0.518 &0.345  & 0.2337  \\

LayoutDM$^{\dagger}$\cite{inoue2023layoutdm} &43M  
        & 0.151 & 0.0161 &0.672 &0.260  & 0.3224 
        & 0.152 & 0.0168 &0.892 &0.778  & 0.0325  \\

RALF\cite{horita2023retrieval} &43M 
        & 0.138 & 0.0126 & 0.977 &0.907  & 0.0069 
        & 0.140 & \textbf{0.0141} & 0.994 & 0.980
        & 0.0046\\
  % \rowcolor{Gray}      
 \textbf{LayoutDiT(Ours)} &49M 
        & \textbf{0.111}& \textbf{0.0125}& \textbf{1.000}& \textbf{0.996}& \textbf{0.0012}
        &\textbf{0.116}&0.0151	& \textbf{0.998}	&\textbf{0.989}	&\textbf{0.0014} \\
        
\bottomrule
\end{tabular}
}
\end{table*}
\begin{table*}[htbp]
\centering
\caption{Comparison of constrained generation methods on PKU and CGL annotated datasets.}
\vspace{-5pt}
\label{tab:con_all_quantitative}
\resizebox{0.9\linewidth}{!}{
\begin{tabular}{lcccccccccc}
% \begin{tabularx}{\textwidth}{@{}Xccccccccccccccc@{}}
\toprule
\quad  & \multicolumn{5}{c}{PKU annotated} & \multicolumn{5}{c}{CGL annotated} \\
\cmidrule(lr){2-6} \cmidrule(lr){7-11}

Method   & \multicolumn{2}{c}{Content} & \multicolumn{3}{c}{Graphic} & \multicolumn{2}{c}{Content} & \multicolumn{3}{c}{Graphic} \\

\cmidrule(lr){2-3} \cmidrule(lr){4-6} \cmidrule(lr){7-8} \cmidrule(lr){9-11}

\quad & Occ~$\downarrow$ & Rea~$\downarrow$ & Und\mysub{L}~$\uparrow$ & Und\mysub{S}~$\uparrow$ & Ove~$\downarrow$ & Occ~$\downarrow$ & Rea~$\downarrow$ & Und\mysub{L}~$\uparrow$ & Und\mysub{S}~$\uparrow$ & Ove~$\downarrow$ \\ 
\midrule

\textbf{C} $\rightarrow$ \textbf{S}  + \textbf{P}\\

LayoutDM$^{\dagger}$ 
                    &0.161&0.0171&0.669&0.274&0.2949&
                    0.141&0.0150&0.924&0.830&0.0332  \\

RALF 
                    &0.136&0.0127&0.967&0.890&0.0091  & 
                    0.140&0.0142&0.987&\textbf{0.969}&\textbf{0.0056}  \\

 \textbf{LayoutDiT(Ours)} &
\textbf{0.121}&\textbf{0.0123}&\textbf{0.989}&\textbf{0.962}& \textbf{0.0025}& \textbf{0.117}&\textbf{0.0137}&\textbf{0.990}&0.959&0.0077 \\

\bottomrule

\textbf{C} $+$ \textbf{S} $\rightarrow$ \textbf{P}\\

LayoutDM$^{\dagger}$  
                &0.162&0.0152&0.755&0.563&0.2063
                &0.142&0.0149&0.922&0.834&0.0268   \\

RALF 
                & 0.138 & 0.0123&0.945&0.876&0.0113
                &0.143&0.0148&0.984&\textbf{0.957}&\textbf{0.0051}   \\

 \textbf{LayoutDiT(Ours)} 
                & \textbf{0.133} & \textbf{0.0122} & \textbf{0.986}& \textbf{0.922} & \textbf{0.0059}
                & \textbf{0.127} & \textbf{0.0131} & \textbf{0.988} & 0.904 & 0.0075 \\

\bottomrule

\textbf{Completion} \\

LayoutDM$^{\dagger}$ 
                    &0.141&0.0148&0.549&0.235&0.1822 
                    &0.142&0.0149&0.877&0.757&0.0247   \\

RALF 
                    & 0.137 & \textbf{0.0132} &0.970& 0.905 & 0.0139 
                    & 0.141 & 0.0144 &\textbf{0.988}&\textbf{0.966}& 0.0042\\

 \textbf{LayoutDiT(Ours)} 
                & \textbf{0.119} & 0.0135 & \textbf{0.981}& \textbf{0.951}  & \textbf{0.0027}
                & \textbf{0.123} & \textbf{0.0142} & 0.978 & 0.933 & \textbf{0.0039}\\

\bottomrule

\textbf{Refinement} \\

LayoutDM$^{\dagger}$  
                    &0.129&0.0112&0.910&0.599&0.0054
                    &0.143&0.0145&0.890&0.614&0.0036  \\

RALF 
                    & 0.128 & 0.0103 &0.989&0.951  & 0.0047
                    & 0.142 & 0.0136 &\textbf{0.994}&\textbf{0.980}& 0.0024\\

 \textbf{LayoutDiT(Ours)} 
                    & \textbf{0.127} & \textbf{0.0094} &\textbf{0.990} & \textbf{0.965}&\textbf{0.0009} 
                    & \textbf{0.140} & \textbf{0.0109} & 0.987 & 0.946
                    & \textbf{0.0019}\\

\bottomrule

\end{tabular}
}
\end{table*}

\noindent \textbf{Quantitative comparison.} 
We present the quantitative comparison on the annotated datasets without constraints in Tab.~\ref{tab:un_all_annotated_quantitative}.
Our method achieves the best results overall, except for the \text{Rea} metric of RALF on CGL.
Notably, our approach demonstrates exceptional performance in the \text{Und} and \text{Ove} metrics, reflecting its effectiveness in modeling the layout graphic structure.
Additionally, as shown in Tab.~\ref{tab:small-sized}, our method effectively reduces the small-sized elements. While LayoutDM achieves 0\% on PKU, it suffers from severe overlap issues. 
These results show our method's balanced excellence in both content and graphic metrics.

\begin{table*}[htbp]

\begin{minipage}[b]{.52\linewidth}
\centering
\resizebox{\linewidth}{!}{
\begin{tabular}{@{}c c c c c c c c c c c c @{}}
\toprule
Train & Test & Method & Occ~$\downarrow$ & Rea~$\downarrow$ & Und\mysub{L}~$\uparrow$ & Und\mysub{S}~$\uparrow$ & Ove~$\downarrow$
\\
\midrule
\multirow{4}{*}{CGL} & \multirow{2}{*}{PKU-unanno} & RALF & 0.143&\textbf{0.0170}&0.989&0.958&0.0347  
\\ 
~ & ~ & Ours
& \textbf{0.115}&0.0178&\textbf{0.998}&\textbf{0.980}& \textbf{0.0025}
\\
\cmidrule{2-8}
~ & \multirow{2}{*}{PKU-anno} &RALF&0.140& \textbf{0.0113}& 0.995&0.976&0.0078 
\\ 
~ & ~ & Ours
& \textbf{0.108}&0.0142&\textbf{0.995}&\textbf{0.987}&\textbf{0.0015}
\\

\midrule
\multirow{4}{*}{PKU} &
\multirow{2}{*}{CGL-unanno} &
RALF &0.342&0.0425&0.927&0.773& 0.0309  
\\ 
~&~& Ours
& \textbf{0.328}&\textbf{0.0331}&\textbf{1.000}&\textbf{0.992} &\textbf{0.0053}
\\
\cmidrule{2-8}
~ & \multirow{2}{*}{CGL-anno} &RALF&0.150&0.0170&0.974&0.902&0.0065
\\ 
~ & ~ & Ours
& \textbf{0.128}&\textbf{0.0147}&\textbf{0.999}&\textbf{0.995}&
\textbf{0.0040}
\\

\bottomrule
\end{tabular}
}
\vspace{1mm}
\captionof{table}{Cross-dataset evaluation setup: We train a model on PKU and test it on CGL, or vice versa.}
\label{tab:cross_uncond}
\end{minipage}\hfill
\begin{minipage}[b]{.45\linewidth}
\centering
\includegraphics[width=\linewidth]{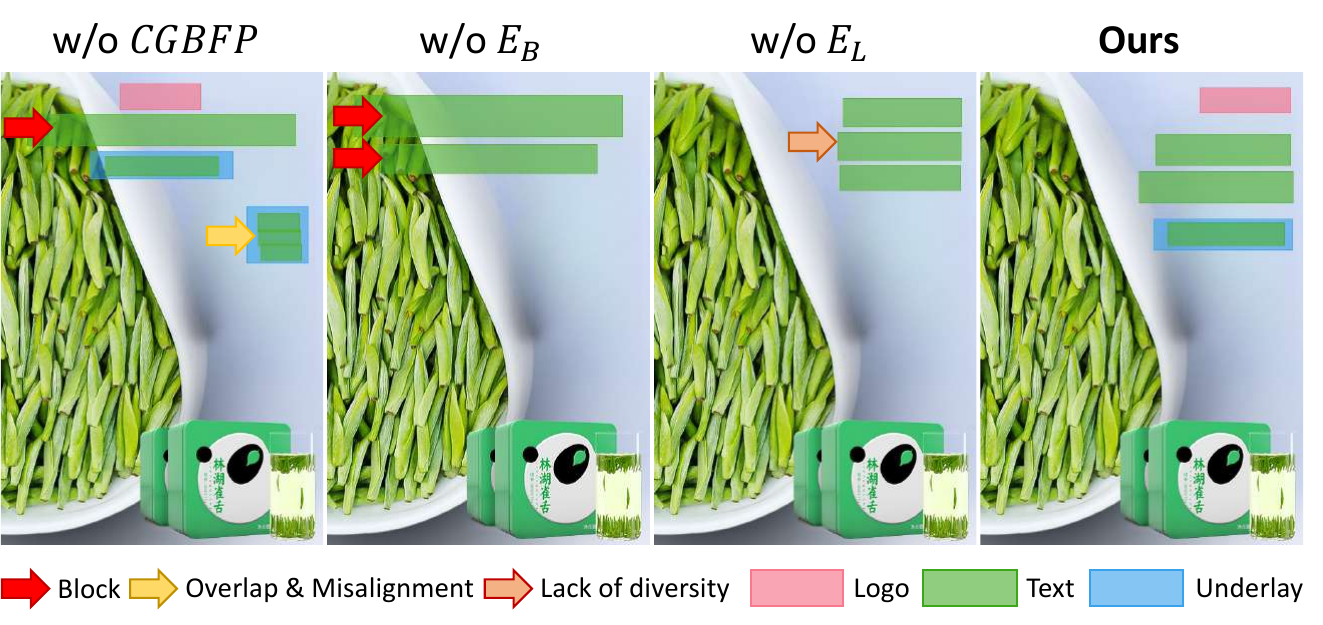}
\captionof{figure}{Examples of ablation studies on module contributions and their effects.}
\label{fig:pku_ablation_uncon}
\end{minipage}
\vspace{-8pt}
\end{table*}
\noindent \textbf{Out-of-domain generalization.}
To further validate the generalization of our method, we conducted cross-dataset evaluation between PKU and CGL. As shown in Tab.~\ref{tab:cross_uncond}, our method outperformed RALF in most metrics. The findings demonstrate our approach adapts well to diverse real-world images, showcasing strong generalization capabilities.

\subsection{Constrained Generation}
\begin{figure*}[ht]
\begin{center}
\centerline{\includegraphics[width=1\linewidth]{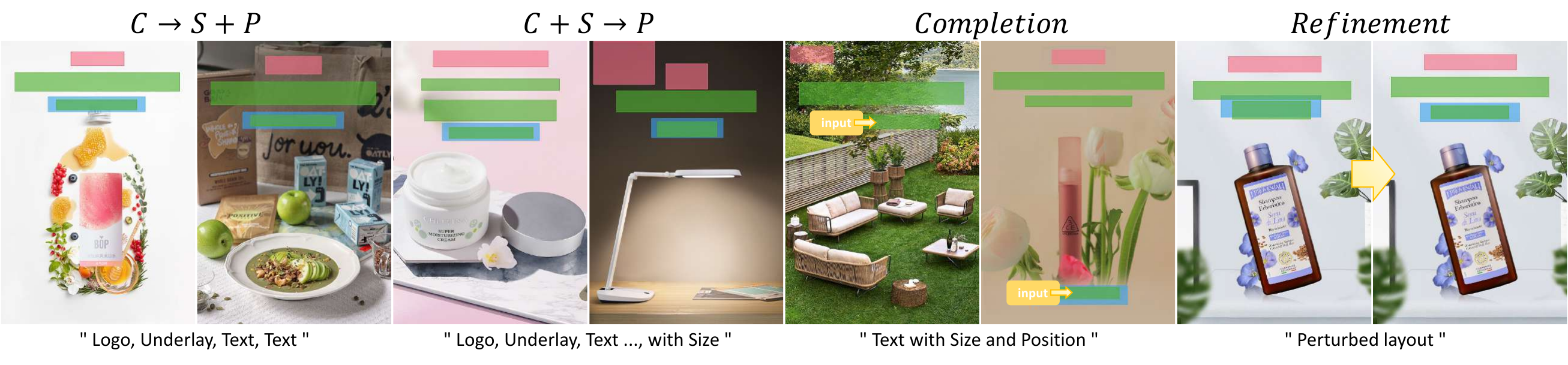}}
\vspace{-2pt}
\caption{Qualitative results of our method in constrained setting on PKU. 
Examples of input constraints and generated results for each constrained generation task. 
Quotation marks indicate the constraints.}
\label{fig:pku_qualitative_con_1}
\end{center}
\vspace{-24pt}
\end{figure*}

Designing constrained experiments allows us to tailor generated outputs to meet specific needs under various conditions. Following the framework established for content-agnostic generation~\cite{jiang2023layoutformer++}, we assess various approaches across constrained tasks within a content-aware generation.

$Category \rightarrow Size + Position (C \rightarrow S + P)$ means we specify the type of layouts, and the model generates the appropriate sizes and positions for each layout. 

$Category + Size \rightarrow Position (C + S \rightarrow   P)$ involves providing the types and sizes of layout elements as input and generating the position of each element accordingly. 

$Completion$ fixes certain element properties randomly and then generates a complete layout.

$Refinement$ aims to correct layouts that have been perturbed with Gaussian noise, with a mean of 0 and a variance of 0.01, added to the true label values following~\cite{rahman2021ruite}.

Benefiting from the properties of diffusion models, our method supports constrained settings without requiring additional constraint modules. Specifically, we only need to treat the polluted data as an intermediate state in the reverse process for the refinement task, eliminating the need for additional training. For other tasks, we train and infer the corresponding constraint conditions in the form of masks.

The generated examples are shown in Fig.~\ref{fig:pku_qualitative_con_1}, with quantitative results reported in Tab.~\ref{tab:con_all_quantitative}. These results demonstrate our method's effectiveness in constrained tasks, achieving strong performance without constraint encoder.

\subsection{Ablation Study}
\begin{table*}[!t]
\centering
\caption{Ablation studies on PKU and CGL annotated datasets.}
\label{tab:concat_ablation_anno}
\resizebox{0.9\linewidth}{!}{
\begin{tabular}{lcccccccccc}
\toprule
\multirow{2}{*}{\makebox[3em][l]{$E_L$}\makebox[3em][l]{$E_B$}\makebox[3em][l]{CGBFP}} & 
\multicolumn{5}{c}{PKU annotated} &\multicolumn{5}{c}{CGL annotated} \\
\cmidrule(lr){2-6} \cmidrule(lr){7-11}
&  Occ$\downarrow$ & Rea$\downarrow$
&  Und\mysub{L}$\uparrow$ & Und\mysub{S}$\uparrow$ & Ove$\downarrow$
&  Occ$\downarrow$ & Rea$\downarrow$
&  Und\mysub{L}$\uparrow$ & Und\mysub{S}$\uparrow$ & Ove$\downarrow$\\
\midrule
\makebox[3em][l]{\checkmark}\makebox[4em][l]{}\makebox[2em][l]{} & 
0.126 & 0.0145& 0.995&0.965 &0.0045&
0.124& 0.0161&0.993 & 0.952&0.0031
\\ 
\makebox[3em][l]{\checkmark}\makebox[4em][l]{\checkmark}\makebox[2em][l]{} & 
0.116 & 0.0130& 0.999 &0.987 &0.0027&
0.120 & 0.0155& 0.997&0.984 &0.0019
\\ 
\makebox[3em][l]{\checkmark}\makebox[4em][l]{}\makebox[2em][l]{\checkmark} & 
 0.115&0.0130&0.998 &0.995 & 0.0021&
 0.118& \textbf{0.0151}& 0.997&0.985 &0.0011
\\ 
\makebox[3em][l]{}\makebox[4em][l]{\checkmark}\makebox[2em][l]{\checkmark} & 
0.123 &0.0146&\textbf{1.000}&0.994&0.0023&
 0.130&0.0165 & \textbf{0.998} &\textbf{0.989} & \textbf{0.0009}
\\
\midrule
\makebox[3em][l]{\checkmark}\makebox[4em][l]{\checkmark}\makebox[2em][l]{\checkmark} &  
\textbf{0.111}& \textbf{0.0125}& \textbf{1.000}& \textbf{0.996}& \textbf{0.0012}&
\textbf{0.116}&\textbf{0.0151}& \textbf{0.998}&\textbf{0.989}	&0.0014 \\
\bottomrule
\end{tabular}
}
\vspace{-10pt}
\end{table*}
We conducted extensive ablation studies to validate the proposed CGBFP, Bounding Encoder($E_B$) and Layout Encoder($E_L$). We provide detailed examples in Fig.~\ref{fig:pku_ablation_uncon} to show the function of each module. Tab.~\ref{tab:concat_ablation_anno} shows the quantitative results on the annotated datasets.

\noindent \textbf{Saliency bounding box.} Comparing the results of the second row to the base model(the first row) demonstrates that employing the saliency bounding box leads to improvements in both content and graphic metrics. Fig.~\ref{fig:pku_ablation_uncon} shows that explicit geometric information extraction helps the model better perceive salient regions, reducing blocking issues.

\noindent \textbf{CGBFP.} The third row shows that CGBFP module enhances overall model performance, significantly improving both content and graphic metrics. Fig.~\ref{fig:pku_ablation_uncon} further shows that removing the factor $\omega$ leads to imbalances between content and graphic, such as occlusion between text elements and salient regions and overlap among text elements.

\noindent \textbf{Layout encoder.} We conducted experiments by removing the layout encoder while maintaining parameter count. Results show decreased content performance without the layout encoder.  More importantly, as shown in Fig.~\ref{fig:pku_ablation_uncon}, generated layouts without a layout encoder lack diversity, particularly in category variety, likely because the encoder effectively decouples the learning of layouts and images.

\vspace{-5pt}
\section{Conclusion}
In this work, we apply a diffusion transformer architecture to content-aware layout generation, which aims to address key challenges in previous work, such as blocking, overlapping, misalignment, and small-sized. 
Yet, directly using the DiT model still faces difficulties in balancing content-aware and graphic features.
This is mainly because the layout generation space plays a crucial role in the modeling process.
Therefore, we introduce a content-graphic balance factor that dynamically modulates layout representations and image features to optimize the model's awareness of the layout generation space.
Additionally, we add the saliency bounding box to bridge the modality gap between images and layouts.
We conduct extensive experiments that demonstrate that our method achieves SOTA performance under both unconstrained and constrained settings. 

\noindent \textbf{Discussions.} 
To our best knowledge, only the PKU and CGL datasets are available as open-source resources. However, given our method's strong generalization ability and capability in modeling diverse layout graphic structures, we believe it can perform well in more complex scenarios, like mobile UIs and shopping websites.
Nevertheless, evaluating performance on more complex scenarios requires specific datasets, which we plan to explore in future work.

\noindent \textbf{Limitations}
We acknowledge two main limitations of this work: 
1) Our framework relies on predefined element categories, which restricts generated layouts to a finite set and limits their applicability to diverse real-world graphic design. Exploring open-vocabulary layout generation remains a necessary future direction.
2) Our current approach primarily uses images as control conditions. We highlight that incorporating additional constraints, such as text-based conditions, represents a promising direction for future research.

% \noindent \textbf{Potential societal impacts.} Positive societal impacts of our CGB-DM include the ability to significantly enhance the efficiency and creativity of graphic design processes. By automating layout generation, designers can focus more on creative tasks, potentially leading to more innovative and visually appealing designs. However, as with other generative models, our CGB-DM could inadvertently create fake advertisements or magazine layouts, potentially leading to deception and the spread of misinformation.

{ 
    \small
    \bibliographystyle{ieeenat_fullname}
    \bibliography{main}
}
\clearpage

\renewcommand\thesection{\Alph{section}}
\renewcommand\thesubsection{\thesection.\arabic{subsection}}
\renewcommand\thefigure{\Alph{section}.\arabic{figure}}
\renewcommand\thetable{\Alph{section}.\arabic{table}} 

\setcounter{section}{0}
\setcounter{figure}{0}
\setcounter{table}{0}

\twocolumn[
\begin{center}
    % \noindent{\Large{\textbf{Appendix}}}
    \Large{\textbf{\title{Supplementary Material for \\ \textit{LayoutDiT: Exploring Content-Graphic Balance in Layout Generation with Diffusion Transformer}}}}
\end{center}
]

% In this Appendix, we provide additional results and analysis.
\section*{Table of contents:}
\begin{itemize}
    \item Section~\ref{sec:A} : Implementation Details
    \item Section~\ref{sec:B} : Additional Experimental Results and Analysis
\end{itemize}

\section{Implementation Details}\label{sec:A}

\subsection{Dataset characteristics}\label{sec:A.1}

We summarize the characteristics of CGL dataset and PKU dataset in Table~\ref{tab:com_dataset}. The CGL dataset contains an additional decorative element for embellishment purposes.
The table presents two key statistical metrics. “Elements per Image” represents the average number of layout elements per image, measuring the degree of layout complexity. “Saliency Mean” represents the proportion of space occupied by the saliency map. We observe a notable disparity between the training and test sets of CGL in “Saliency Mean”. This can be attributed to the test sets of CGL being characterized by complex backgrounds and large saliency regions in the images. 

\begin{table}[htb]
\setlength{\heavyrulewidth}{2.5pt}
\huge
\centering
\caption{Statistical characteristics of PKU and CGL datasets.}
\resizebox{\linewidth}{!}{
\renewcommand{\arraystretch}{1.5}
\begin{tabular}{cccccccc}
% \hline
\toprule
\multirow{2}{*}{Dataset} &
  \multicolumn{2}{c}{Dataset Size} &
  \multirow{2}{*}{Elements} &
  \multirow{2}{*}{Categories} &
  \multirow{2}{*}{\begin{tabular}[c]{@{}c@{}}Elements\\per Image\end{tabular}} &
  \multicolumn{2}{c}{Saliency Mean} 
  \\ \cline{2-3} \cline{7-8}
 & Train & Test & & & & Train & Test \\ 
\midrule
CGL &
  60,548 &
  1,000 &
  \begin{tabular}[c]{@{}c@{}}Text, logo,\\underlay, embel.\end{tabular} &
  \begin{tabular}[c]{@{}c@{}}Cosmetics, electronics,\\clothing\end{tabular} &
  4.715 &
  0.383 &
  0.459 \\
\midrule
% \hline
PKU &
  9,974 &
  905 &
  \begin{tabular}[c]{@{}c@{}}Text, logo,\\underlay\end{tabular} &
  \begin{tabular}[c]{@{}l@{}}Cosmetics, electronics,\\clothing, delicatessen,\\toys/instruments\end{tabular} &
  4.878 &
  0.378 &
  0.351 \\
\bottomrule
\end{tabular}
}
\label{tab:com_dataset}
\end{table}

Finally, we need to point out the data distribution differences in the datasets we use.
These differences manifest in two aspects. First, PKU and CGL datasets differ significantly in their element types, layout complexity, layout distribution, and content categories, as noted in DS-GAN~\cite{hsu2023posterlayout} Section 3. Second, distribution gaps exist between training and test sets within each dataset, primarily due to varying image types and the domain gap between inpainted and clean canvases. Due to these distribution differences, conducting comprehensive experiments and cross-dataset evaluation (Section 4.3 in the main text) helps assess the generalization ability and robustness of the methods.

\subsection{Data processing}\label{sec:A.2}
\begin{figure}[ht]
\begin{center}\tiny
\centerline{\includegraphics[width=1\linewidth]{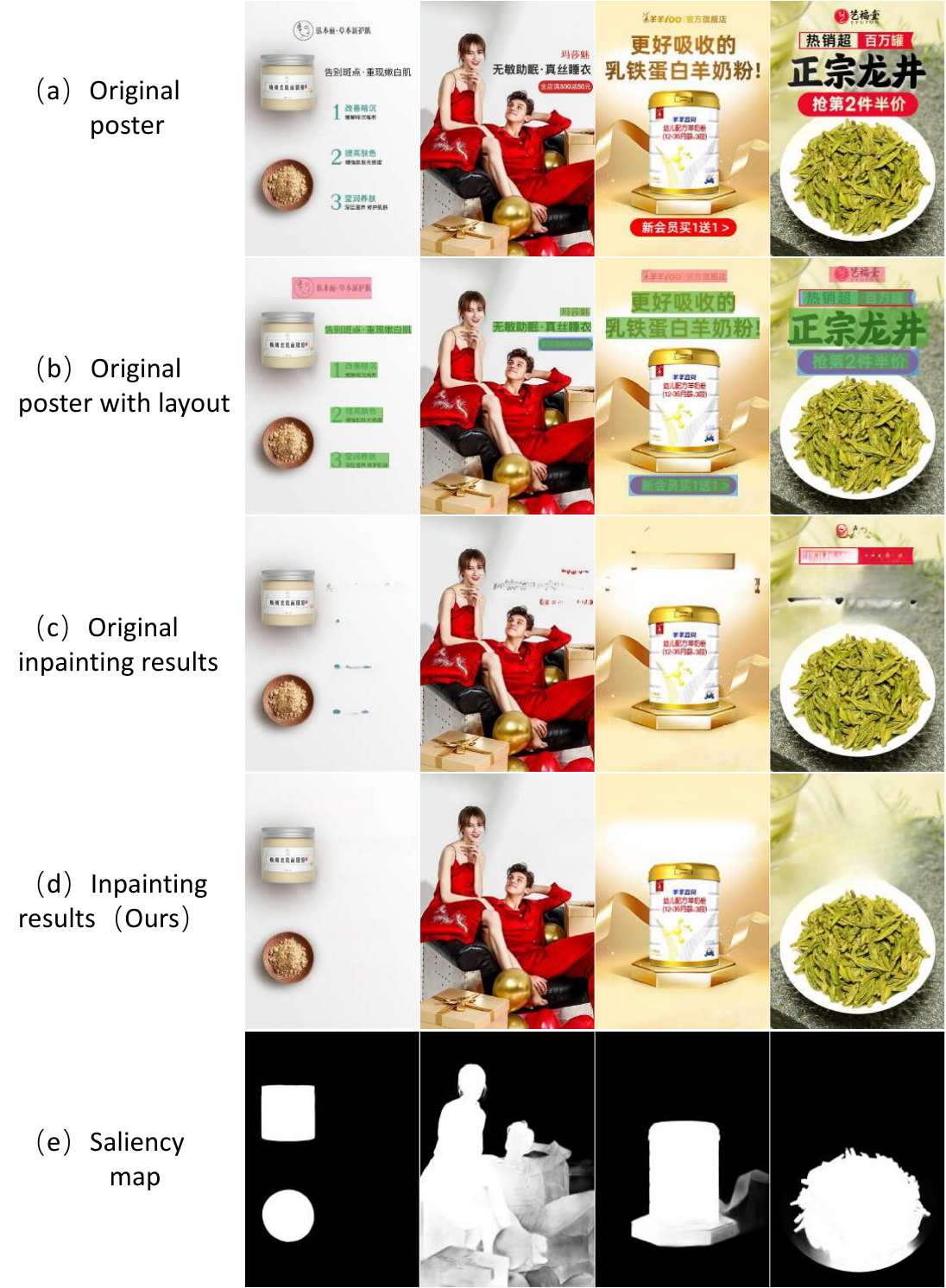}}
\vspace{-2pt}
\caption{Comparison of inpainting for the dataset preprocessing.}
\label{fig:inpaint}
\end{center}
\vspace{-8pt}
\end{figure}

To obtain canvas-layout pairs for training, we need to inpaint images containing visual elements. While PKU provides inpainted images in their dataset, CGL only offers images with authentic labels. However, the quality of PKU's inpainted images is suboptimal, as they retain visible traces of the original layout elements. We argue that these residual traces could adversely affect the model training process. Therefore, to ensure a fair comparison and obtain high-quality training data, we opt  to employ the inpainting algorithm \cite{suvorov2022resolution} to enhance the repaired regions in both datasets. Fig.~\ref{fig:inpaint} illustrates specific examples of this process.

Saliency map is essential for providing comprehensive spatial information of images. While previous methods mainly rely on the PFPN~\cite{wang2020progressive} and BASNet~\cite{qin2019basnet} for saliency map extraction, the connectivity of PFPN's weights has become defunct. Consequently, we resort to ISNet~\cite{qin2022highly} and BASNet for reobtaining saliency maps. In line with previous methods, we combine the saliency maps obtained from both algorithms using a maximum operation. Fig.~\ref{fig:inpaint} (e) presents the saliency maps generated by ISNet.

To obtain the annotated test set, we follow RALF's division criteria, splitting the original training dataset with a train/validation/test ratio of 8:1:1. Similarly, we exclude samples with more than 11 elements.

\subsection{Architecture details}\label{sec:A.3}

As shown in Table~\ref{tab:network parameters}, we provide all the details of the architecture.

\begin{table}[htb]\footnotesize
	\centering
	\caption{Network parameters.}
	\begin{tabular}{ccccc}
		\toprule
		Module & Layers & Dim & FFN-dim & Head \\
		\midrule
		Layout Encoder & 2 & 512 & 1024 & 8 \\
		Layout Decoder & 4 & 512 & 1024 & 8 \\
		Image Encoder & 6 & 512 & 2048 & 8 \\
		CGBFP& 2 & 512 & 2048 & 8\\
		\bottomrule
	\end{tabular} 
 \label{tab:network parameters}
\end{table}
\noindent \textbf{Layout Encoder $\&$ Layout Decoder.}
Both the layout encoder and layout decoder consist of a series of transformer blocks. We use a two-layer transformer in the encoder and a four-layer transformer in the decoder. The key difference is that the layout decoder incorporates a cross-attention module to interact with the image features and saliency bounding box features.

\noindent \textbf{Image Encoder.}
The image encoder consists of ViT-like transformer blocks for effective image features extraction. We resize the input images to $384\times256$ to match the resolution of the original data, using a patch size of 32.

\noindent \textbf{CGBFP.}
The CGBFP module is implemented using a trainable query transformer network. The architecture consists of a two-layer q-former network that takes N learnable query embeddings as input, which interact with image embeddings through attention layers. Softplus is selected as the output activation function.

\noindent \textbf{Bounding Encoder.}
We utilize a bounding encoder implemented with a three-layer MLP network to encode the salient bounding box, matching the dimensions required by the subsequent model. Specifically, the network's input dimension is 4, while both the intermediate and output dimensions are 512. Softplus activation function is used throughout.

\subsection{User study details}\label{sec:A.4}

We provide a detailed introduction about our user study. We randomly select 20 images as test cases. To eliminate potential bias, we present the results from different methods in randomized order for each question. We recruit 25 participants with normal vision to evaluate the generated results. The participants are asked to assess the eligibility of the generated layouts and identify the most preferred option for each image. We quantify the results using two metrics: the selection rate ($P_{sel}$) and preference rate ($P_{pref}$). 

\noindent \textbf{Instructions.}
\begin{itemize}

\item \textbf{Task.} The layout generation task aims to generate appropriate layouts for given e-commerce poster images, which are intended for rendering applications. In this user study, you will evaluate layouts generated by 6 different models across 20 sample cases. For each case, you are required to complete two evaluation tasks:
\begin{itemize}
\item Layout Validity Assessment (Multiple Choice): Evaluate whether each model's generated layout meets the quality criteria
\item Optimal Layout Selection (Single Choice): Select the most satisfactory layout from the six generated results
\end{itemize}

\item \textbf{Element Taxonomy.}
The layouts comprise four distinct element categories:
\begin{itemize}
\item \textbf{Logo}: Brand identifiers serving information transmission purposes
\item \textbf{Text}: Textual containers for information conveyance
\item \textbf{Underlay}: Background elements that can be positioned beneath any non-underlay element, enhancing text readability and mitigating complex background interference
\item \textbf{Embellishment}: Decorative elements (exclusively present in the CGL dataset with lower frequency)
\end{itemize}

\item \textbf{Quality Assessment Criteria.} The generated layouts should show well-structured graphic design and
seamlessly integrates with the image content. The evaluation criteria are categorized as follows:
\begin{itemize}

\item Image-Level Criteria
\begin{itemize}
\item Logo and text elements should minimize occlusion of primary poster regions
\item Underlay elements may overlap with image content, serving as background coverage
\item Generated layouts should achieve seamless integration with the image content, maintaining visual harmony between layout elements and the poster
\end{itemize}
\item Layout-Level Criteria
\begin{itemize}
\item Non-underlay elements (logo, text) should avoid mutual overlap
\item Underlay elements should optimally encompass their associated non-underlay elements
\item Generated layouts should demonstrate well-structured graphic design principles that align with human aesthetic preferences
\end{itemize}

\end{itemize}

\end{itemize}

\subsection{Constrained generation details}\label{sec:A.5}
We eliminate the need for additional constraint encoders commonly used in other methods to encode conditions in constrained generation experiments. Leveraging the advantages of diffusion models, we only need to modify the details in the forward diffusion and sampling processes. Specifically, for tasks including $C \rightarrow S + P$, $C + S \rightarrow P$, and $Completion$, we simply add masks for known attributes during training to preserve their ground truth values without noise corruption. The same principle applies to the inference process. We observe that training separate models for individual tasks yields superior performance, as shown in Section 4.5 Constrained Generation. However, using a single model to handle all tasks simultaneously is feasible and represents a promising direction for future research. For the $Refinement$ task, no additional training is required as we can directly utilize the model trained for unconditioned generation. We only need to treat the polluted data as an intermediate state in the reverse process.

Given that the model has access to more conditional information during optimization, we modify several parameters. Specifically, we adjust the number of training epochs to 400 and change the time-step sampling schedule from “uniform” to “quad”.

\section{Additional Experimental Results and Analysis}\label{sec:B}
We provide additional quantitative results and qualitative comparisons on both the PKU and CGL datasets. 

\subsection{Unconstrained generation on unannotated datasets}\label{sec:B.1}

\begin{table}[htbp]
\centering
\caption{Comparison of unconstrained generation methods on PKU unannotated datasets.}
\label{tab:pku_unanno_unconstrain_appendix}
\resizebox{\linewidth}{!}{
\begin{tabular}{@{}ccccccccc@{}}
% \begin{tabularx}{\textwidth}{@{}Xccccccccccccccc@{}}
\toprule
\quad & \quad & \multicolumn{6}{c}{PKU unannotated} \\
\cmidrule{3-8} 
Method & Params   & \multicolumn{2}{c}{Content} & \quad & \multicolumn{3}{c}{Graphic} \\
\cmidrule{3-4} \cmidrule{6-8} 

\quad & \quad  & Occ~$\downarrow$ & Rea~$\downarrow$ & \quad & Und\mysub{L}~$\uparrow$ & Und\mysub{S}~$\uparrow$ & Ove~$\downarrow$ \\ 
\midrule
CGL-GAN\cite{zhou2022composition} &41M 
        & 0.213 & 0.0256 &&0.718& 0.299  & 0.1034  \\

DS-GAN\cite{hsu2023posterlayout}  &30M    
        & 0.179 & 0.0228 && 0.764  & 0.464& 0.0368 \\

ICVT\cite{cao2022geometry}  &50M      
        & 0.295 & 0.0254 &&0.420 &0.301  & 0.3167  \\

LayoutDM$^{\dagger}$\cite{inoue2023layoutdm} &43M  
        & 0.153 & 0.0220 &&0.645 &0.235  & 0.3051  \\

RALF\cite{horita2023retrieval} &43M 
        & 0.142 & 0.0190 && 0.965 &0.878  & 0.0106 \\
        
\textbf{LayouDiT(Ours)} &49M 
        & \textbf{0.124}&\textbf{0.0161}&&\textbf{1.000}&\textbf{0.994}&\textbf{0.0004} \\
\bottomrule
\end{tabular}
    }
\end{table}

Table~\ref{tab:pku_unanno_unconstrain_appendix} demonstrates that even on the unannotated PKU dataset, our method achieves significant advantages. Specifically, regarding underlay effectiveness and overlap issues, our method performs comparably to its performance on the annotated dataset. In contrast, other baselines exhibit some degree of performance degradation. This indicates that our method can better overcome data distribution differences and possesses superior generalization capabilities.

\begin{table}[htbp]

\centering
\caption{Comparison of unconstrained generation methods on CGL unannotated datasets.}
\label{tab:cgl_unanno_unconstrain_appendix}
\resizebox{\linewidth}{!}{
\begin{tabular}{@{}cccccccccc@{}}
% \begin{tabularx}{\textwidth}{@{}Xccccccccccccccc@{}}
\toprule
\quad & \quad & \multicolumn{7}{c}{CGL unannotated} \\
\cmidrule{3-9} 
Method & Params   & \multicolumn{3}{c}{Content} & \quad & \multicolumn{3}{c}{Graphic} \\
\cmidrule{3-5} \cmidrule{7-9} 

\quad & \quad  & Uti~$\uparrow $ & Occ~$\downarrow$ & Rea~$\downarrow$ & \quad & Und\mysub{L}~$\uparrow$ & Und\mysub{S}~$\uparrow$ & Ove~$\downarrow$ \\ 
\midrule
CGL-GAN\cite{zhou2022composition} &41M  & 0.072
        & 0.496 & 0.0604 &&0.652& 0.129  & 0.2478  \\

DS-GAN\cite{hsu2023posterlayout}  &30M   & \textbf{0.131}
        & 0.407 & 0.0533 && 0.804  & 0.349& 0.0831  \\

ICVT\cite{cao2022geometry}  &50M      & 0.058
        & 0.464 & 0.0502 &&0.466 &0.306  & 0.1959  \\

LayoutDM$^{\dagger}$\cite{inoue2023layoutdm} & 43M  & 0.024
        & 0.432 & 0.0499 &&0.730 &0.539  & 0.0979  \\

RALF\cite{horita2023retrieval} &43M & 0.056
        & \textbf{0.258} & \textbf{0.0319} && 0.986 & 0.935& 0.0572\\

\textbf{LayoutDiT(Ours)} &49M & 0.106
        & 0.339&0.0358 &&\textbf{0.998} & \textbf{0.989}& \textbf{0.0023}& 	 \\
\bottomrule
\end{tabular}
}
\end{table}

We also conduct comparative tests on the unannotated CGL dataset, with results presented in Table~\ref{tab:cgl_unanno_unconstrain_appendix}. All methods show performance degradation on this test set.
Firstly, regarding graphic metrics, baseline methods show substantial performance degradation compared to their results on the annotated dataset. In contrast, our method demonstrates more robust performance, achieving scores of 0.0023 on \text{Ove}, 0.998 on $\text{Und}_\text{L}$, and 0.989 on $\text{Und}_{\text{S}}$. Notably, other methods' \text{Ove} scores exceed 0.05, substantially impacting layout quality.

Furthermore, in content metrics, our method performs inferior to RALF but surpasses other baselines. To analyze this phenomenon, we introduced the \text{Uti} metric, which measures layout occupancy in non-significant regions. While \text{Uti} cannot serve as a reliable indicator of layout quality (since randomly generated cluttered layouts can also exhibit high \text{Uti}), a low value indicates the model's tendency to generate small-sized elements. As shown in Table~\ref{tab:cgl_unanno_unconstrain_appendix}, only our LayoutDiT and DS-GAN achieve a \text{Uti} above 0.100, while RALF's \text{Uti} is merely 0.056. The observation is further supported by the results of Tab. 1 (in the main text), which reveals that most methods, except ours and CGL-GAN, suffer from small-sized issues on CGL unannotated tests. While generating small-sized elements may improve content metrics, it compromises overall layout quality. As evidenced in Fig.~\ref{fig:cgl_un_un}, despite RALF's strong content metrics, its results show significant content and graphic imbalance due to the small-sized element generation.

Finally, we analyze this phenomenon from the perspective of layout generation space. As shown in Table~\ref{tab:com_dataset}, CGL test sets exhibit significantly different data distributions from training sets, featuring complex backgrounds and large saliency regions. These characteristics constrain the layout generation space, making it difficult for the model to find suitable generation positions. As a result, the model tends to generate small-sized elements.

\begin{table}[htbp]
% \vspace{-6pt}
\centering
\caption{Ablation studies of the factor on PKU and CGL annotated datasets.}
\label{tab:factor_anno}
\renewcommand{\arraystretch}{1.1} % 调整行距
\resizebox{\linewidth}{!}{
\begin{tabular}{ccccccccccc}
\toprule
\quad  & \multicolumn{5}{c}{PKU annotated} & \multicolumn{5}{c}{CGL annotated} \\
\cmidrule(lr){2-6} \cmidrule(lr){7-11}
Factor $\omega$  & \multicolumn{2}{c}{Content} & \multicolumn{3}{c}{Graphic} & \multicolumn{2}{c}{Content} & \multicolumn{3}{c}{Graphic} \\
\cmidrule(lr){2-3} \cmidrule(lr){4-6} \cmidrule(lr){7-8} \cmidrule(lr){9-11}

\quad   & Occ~$\downarrow$ & Rea~$\downarrow$  & Und\mysub{L}~$\uparrow$ & Und\mysub{S}~$\uparrow$ & Ove~$\downarrow$ & Occ~$\downarrow$ & Rea~$\downarrow$ & Und\mysub{L}~$\uparrow$ & Und\mysub{S}~$\uparrow$ & Ove~$\downarrow$ \\ 
\midrule

$\omega$ = 2
        & 0.130  & 0.0140 & \underline{0.999}&0.978 &0.0052 
        & \textbf{0.109} & \textbf{0.0142} & 0.988  & 0.950& 0.0088 \\

$\omega$ = 1
        & \underline{0.116} & \underline{0.0134}& \underline{0.999}&0.987 &0.0027
        & 0.120& 0.0155& 0.997 & 0.984&0.0019  \\

$\omega$ = 0.5
        & 0.129 & 0.0148 &\underline{0.999} &0.986  & 0.0013 
        & 0.132 & 0.0167 &0.998 &\underline{0.992} & 0.0008  \\

$\omega$ = 0.1
        & 0.138 & 0.0165 & 0.997 &0.985  & \underline{0.0002}
        & 0.174 & 0.0207 & \underline{0.999} & \textbf{0.996} & 7e-5\\

$\omega$ = 0.05
        & 0.143 & 0.0162 & 0.997 &0.987  & \textbf{0.0001}
        & 0.181 & 0.0213 & \underline{0.999} & \textbf{0.996} & \underline{6e-5} \\

$\omega$ = 0.01
        & 0.150 & 0.0170 & 0.997 &\underline{0.992}  & \textbf{0.0001}
        & 0.185 & 0.0218 &\textbf{1.000} & \textbf{0.996} & \textbf{4e-5}\\
        
\textbf{Ours}  
        & \textbf{0.111}& \textbf{0.0125}& \textbf{1.000}& \textbf{0.996}& 0.0012
        & \underline{0.116}&\underline{0.0151} & 0.998 &0.989 &0.0014 \\
\bottomrule
\end{tabular}
}
\end{table}

\begin{table}[htbp]
% \vspace{-6pt}
\centering
\caption{Ablation studies of the factor on PKU and CGL unannotated datasets.}
\label{tab:factor_unanno}
\renewcommand{\arraystretch}{1.1} % 调整行距
\resizebox{\linewidth}{!}{
\begin{tabular}{ccccccccccc}
\toprule
\quad  & \multicolumn{5}{c}{PKU unannotated} & \multicolumn{5}{c}{CGL unannotated} \\
\cmidrule(lr){2-6} \cmidrule(lr){7-11}
Factor $\omega$  & \multicolumn{2}{c}{Content} & \multicolumn{3}{c}{Graphic} & \multicolumn{2}{c}{Content} & \multicolumn{3}{c}{Graphic} \\
\cmidrule(lr){2-3} \cmidrule(lr){4-6} \cmidrule(lr){7-8} \cmidrule(lr){9-11}

\quad   & Occ~$\downarrow$ & Rea~$\downarrow$  & Und\mysub{L}~$\uparrow$ & Und\mysub{S}~$\uparrow$ & Ove~$\downarrow$ & Occ~$\downarrow$ & Rea~$\downarrow$ & Und\mysub{L}~$\uparrow$ & Und\mysub{S}~$\uparrow$ & Ove~$\downarrow$ \\ 
\midrule

$\omega$ = 2
        & 0.141& \underline{0.0172} &0.998 &0.973&0.0040 
        &\textbf{0.311}&\textbf{0.0343}&0.972&0.922&0.0280 \\

$\omega$ = 1
        &\underline{0.125}&0.0177&\underline{0.999}&0.989&0.0021
        & 0.346&0.0362&\underline{0.995}&0.980 &0.0042 \\

$\omega$ = 0.5
        & 0.139 & 0.0185 &\underline{0.999} &\textbf{0.995}  & 0.0007 
        & 0.369&0.0376&\textbf{0.998}&0.988&0.0010 \\

$\omega$ = 0.1
        & 0.157 & 0.0198 & 0.997 &\textbf{0.995}  & \underline{0.0001} 
       &0.387&0.0405&\textbf{0.998}&\textbf{0.995}&0.0001\\

$\omega$ = 0.05
        & 0.162 & 0.0202 & 0.997 &0.990  & \textbf{2e-5}
        & 0.392&0.0409&\textbf{0.998}&\underline{0.993}&\underline{5e-5}\\

$\omega$ = 0.01
        & 0.162 & 0.0203 & 0.995 &0.991  & \textbf{2e-5}
        &0.398&0.0411&\textbf{0.998}&\textbf{0.995}&\textbf{4e-5}\\
        
\textbf{Ours}  
&\textbf{0.124}&\textbf{0.0161}&\textbf{1.000}&\underline{0.994}&0.0004
        & \underline{0.339}&\underline{0.0358}&\textbf{0.998} & 0.989& 0.0023 \\
\bottomrule
\end{tabular}
}
\end{table}

\begin{table}
\centering
\caption{Ablation studies on PKU and CGL unannotated datasets.}
\label{tab:concat_ablation_unanno}
\resizebox{\linewidth}{!}{
\begin{tabular}{lcccccccccc}
\toprule
\multirow{2}{*}{\makebox[3em][l]{$E_L$}\makebox[3em][l]{$E_B$}\makebox[3em][l]{CGBFP}} & 
\multicolumn{5}{c}{PKU unannotated} &\multicolumn{5}{c}{CGL unannotated} \\
\cmidrule(lr){2-6} \cmidrule(lr){7-11}
&  Occ$\downarrow$ & Rea$\downarrow$
&  Und\mysub{L}$\uparrow$ & Und\mysub{S}$\uparrow$ & Ove$\downarrow$
&  Occ$\downarrow$ & Rea$\downarrow$
&  Und\mysub{L}$\uparrow$ & Und\mysub{S}$\uparrow$ & Ove$\downarrow$\\
\midrule
\makebox[3em][l]{\checkmark}\makebox[4em][l]{}\makebox[2em][l]{} & 
0.139&0.0192&0.997&0.972&0.0047&
0.350&0.0394&0.987&0.949&0.0042
\\ 
\makebox[3em][l]{\checkmark}\makebox[4em][l]{\checkmark}\makebox[2em][l]{} & 
0.125&0.0177&0.999&0.989&0.0021&
0.346&0.0362 &0.995 &0.980 &0.0042
\\ 
\makebox[3em][l]{\checkmark}\makebox[4em][l]{}\makebox[2em][l]{\checkmark} & 
\textbf{0.124}& 0.0169&\textbf{1.000} & 0.993& 0.0018&
\textbf{0.339}& 0.0361&0.995 & 0.982&0.0028
\\ 
\makebox[3em][l]{}\makebox[4em][l]{\checkmark}\makebox[2em][l]{\checkmark} & 
0.146& 0.0188& 0.998&0.989 &0.0010 &  
0.362&0.0383 &0.997 &0.978 &0.0037

\\
\midrule
\makebox[3em][l]{\checkmark}\makebox[4em][l]{\checkmark}\makebox[2em][l]{\checkmark} &  
\textbf{0.124}&\textbf{0.0161}&\textbf{1.000}&\textbf{0.994}&\textbf{0.0004}&
\textbf{0.339}&\textbf{0.0358} &\textbf{0.998} & \textbf{0.989}& \textbf{0.0023} \\
\bottomrule
\end{tabular}
}

\end{table}
\begin{figure}[ht]
\begin{center}\tiny
\centerline{\includegraphics[width=1\linewidth]{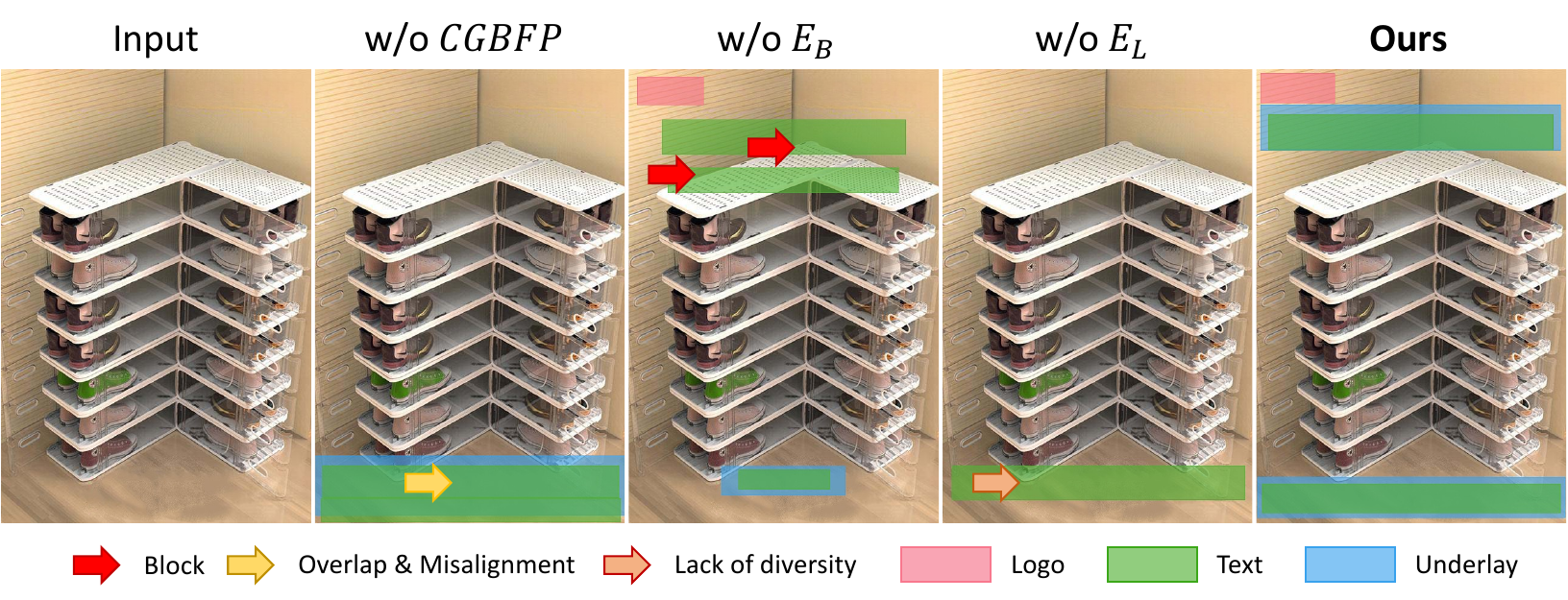}}
\vspace{-2pt}
\caption{Examples of ablation studies on module contributions and their effects.}
\label{fig:pku_ablation_uncon_0}
\end{center}
\vspace{-8pt}
\end{figure}

\subsection{More results on ablation study}\label{sec:B.2}
\begin{figure*}[ht]
\centering
\includegraphics[width=0.8\linewidth]{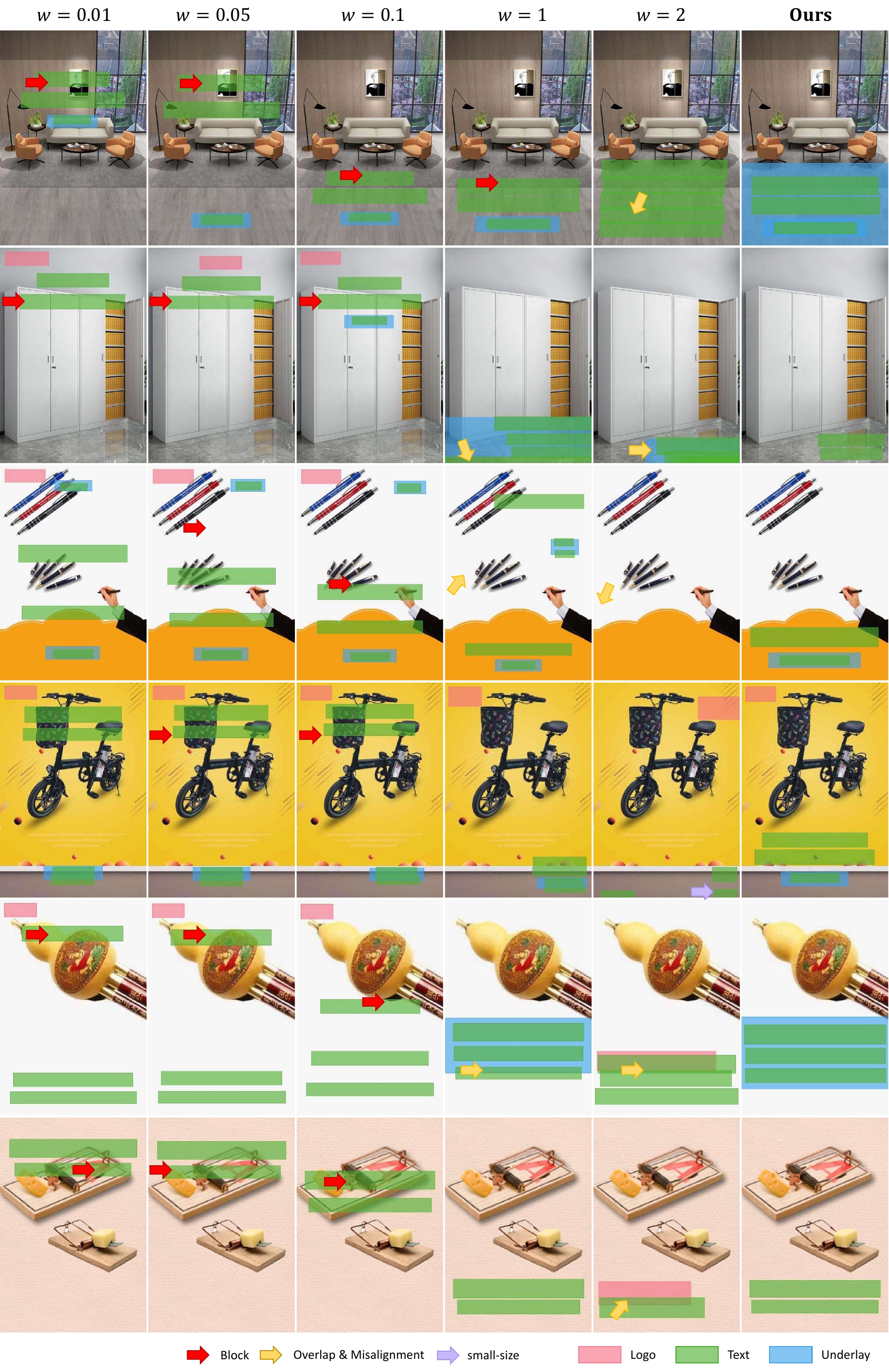}
\caption{Ablation studies of the factor $\omega$. The factor $\omega$ enhances the model’s awareness of layout generation space, thereby
achieving optimal content-graphic balance. In contrast, using a constant factor often leads to poor layout performance in either aspect.}
\label{fig:pku_weight_uncon_appendix}
\end{figure*}

Fig.~\ref{fig:pku_weight_uncon_appendix} presents visual comparisons with different constant values of $\omega$. To demonstrate the advantages of our adaptive factor more intuitively, Table~\ref{tab:factor_anno} and~\ref{tab:factor_unanno} show ablation results of $\omega$ on annotated and unannotated datasets, respectively.

The experimental results reveal that larger $\omega$ values lead to superior content metrics but underperform in graphic metrics. Conversely, smaller values show the opposite trend, indicating the model effectively captures layout graphic structure but overlooks image content perception. Overall, by predicting an adaptive factor through the q-former module, we optimize the model's awareness of the layout generation space, achieving significant improvements in both content and graphic aspects for more comprehensive performance.

We also conduct module ablation experiments on the unannotated test sets. Table~\ref{tab:concat_ablation_unanno} summarizes the experimental results. Additionally, we provide further visual comparisons of the ablation experiments in Fig.~\ref{fig:pku_ablation_uncon_0}.

\begin{table}[htbp]
\centering
\caption{Quantitative evaluation of our method using reduced sampling steps on PKU and CGL annotated datasets.}
\label{tab:step_anno}
\resizebox{\linewidth}{!}{
\begin{tabular}{ccccccccccc}
\toprule
\quad  & \multicolumn{5}{c}{PKU annotated} & \multicolumn{5}{c}{CGL annotated} \\
\cmidrule(lr){2-6} \cmidrule(lr){7-11}
Method  & \multicolumn{2}{c}{Content} & \multicolumn{3}{c}{Graphic} & \multicolumn{2}{c}{Content} & \multicolumn{3}{c}{Graphic} \\
\cmidrule(lr){2-3} \cmidrule(lr){4-6} \cmidrule(lr){7-8} \cmidrule(lr){9-11}

\quad   & Occ~$\downarrow$ & Rea~$\downarrow$  & Und\mysub{L}~$\uparrow$ & Und\mysub{S}~$\uparrow$ & Ove~$\downarrow$ & Occ~$\downarrow$ & Rea~$\downarrow$ & Und\mysub{L}~$\uparrow$ & Und\mysub{S}~$\uparrow$ & Ove~$\downarrow$ \\ 
\midrule

Ours(100)
        & \textbf{0.111}& \textbf{0.0125}& \textbf{1.000}& \textbf{0.996}& \textbf{0.0012}
        &  \textbf{0.116}&\underline{0.0151} & \textbf{0.998} &\textbf{0.989} &\textbf{0.0014} \\

Ours(50)
        & 0.117 &0.0140&0.998 &0.992 &\underline{0.0041}
        &\underline{0.124}& 0.0157&0.993 &0.978&0.0039 \\

Ours(25)
        & \underline{0.115}&0.0137  &\underline{0.999} &\underline{0.993} &0.0043
        &\underline{0.124}& 0.0158&\underline{0.997} &\underline{0.984} &\underline{0.0024} \\

RALF
        & 0.138 & \underline{0.0126} & 0.977 & 0.907 & 0.0069
        & 0.140 & \textbf{0.0141} & 0.994 & 0.980 & 0.0046\\

\bottomrule
\end{tabular}
}
\end{table}

\begin{table}[htbp]
\centering
\caption{Quantitative evaluation of our method using reduced sampling steps on PKU and CGL unannotated datasets.}
\label{tab:step_unanno}
\resizebox{\linewidth}{!}{
\begin{tabular}{ccccccccccc}
\toprule
\quad  & \multicolumn{5}{c}{PKU unannotated} & \multicolumn{5}{c}{CGL unannotated} \\
\cmidrule(lr){2-6} \cmidrule(lr){7-11}
Method  & \multicolumn{2}{c}{Content} & \multicolumn{3}{c}{Graphic} & \multicolumn{2}{c}{Content} & \multicolumn{3}{c}{Graphic} \\
\cmidrule(lr){2-3} \cmidrule(lr){4-6} \cmidrule(lr){7-8} \cmidrule(lr){9-11}

\quad   & Occ~$\downarrow$ & Rea~$\downarrow$  & Und\mysub{L}~$\uparrow$ & Und\mysub{S}~$\uparrow$ & Ove~$\downarrow$ & Occ~$\downarrow$ & Rea~$\downarrow$ & Und\mysub{L}~$\uparrow$ & Und\mysub{S}~$\uparrow$ & Ove~$\downarrow$ \\ 
\midrule

Ours(100)
        &\underline{0.124}&\textbf{0.0161}&\textbf{1.000}&\underline{0.994}&\textbf{0.0004} &
        \underline{0.339}&\underline{0.0358} &\textbf{0.998} & \textbf{0.989}& \textbf{0.0023} \\

Ours(50)
        & \textbf{0.122} & \underline{0.0177} & \textbf{1.000} &\textbf{0.997}& \underline{0.0005}&
        0.350 &0.0381& \underline{0.992}&0.967&0.0046 \\

Ours(25)
        &0.125 & 0.0178 &\underline{0.998} &0.991 &0.0006  &
        0.350&0.0382 &0.991 &\underline{0.972} &\underline{0.0042}\\

RALF
        & 0.142 & 0.0190 & 0.965 &0.878  & 0.0106
        & \textbf{0.258} & \textbf{0.0319} & 0.986 & 0.935& 0.0572\\

\bottomrule
\end{tabular}
}
\end{table}

\subsection{Quantitative results of reduced sampling steps}\label{sec:B.3}

In Table~\ref{tab:step_anno} and~\ref{tab:step_unanno}, we present the quantitative evaluation of our method using reduced sampling steps.
We find that when the sampling steps are within 50 steps, the quadratic time-step sampling schedule yields superior performance.
The results shown in the tables indicate that our method consistently outperforms RALF and achieves state-of-the-art performance even with reduced sampling steps. 
Notably, when the sampling steps are reduced to 25, our method achieves an optimal balance between performance and computational efficiency.

% \input{tex/table/rebuttal/run_time}
% \subsection{Computational cost discussion}\label{sec:B.4}
% For a fair comparison, we conducted runtime and memory usage evaluations on the PKU annotated test set (1000 samples) using a single NVIDIA RTX 4090 GPU. All methods were executed three times and the average results was reported, as shown in Tab.~\ref{tab:run_time}.
% 
% Due to the multiple sampling nature of diffusion models, they do not have an advantage in inference time compared to GAN-based and VAE-based methods. 
%  
%  However, we found that using fewer sampling steps (e.g., 25 steps) can achieve a more comprehensive balance between performance and computational efficiency. 
%  
%  Furthermore, given the low dimensionality of layout data, all methods can generate a set of layouts in approximately one second. 
%  
%  Finally, we believe that employing current advanced diffusion acceleration techniques could further enhance performance.

\subsection{Visual comparison}\label{sec:B.5}
\begin{figure*}[ht]
\begin{center}
\centerline{\includegraphics[width=0.8\linewidth]{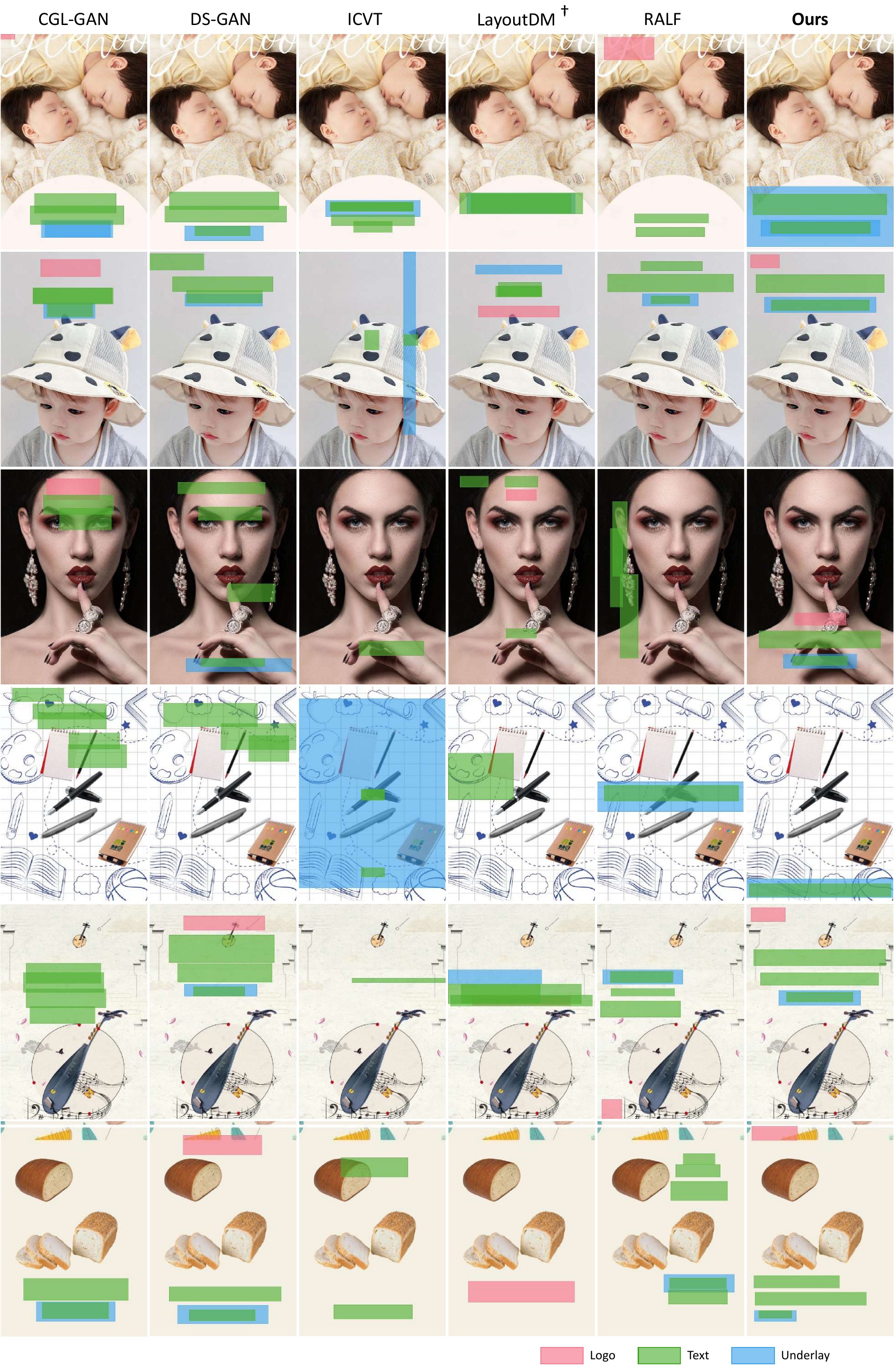}}
\vspace{-2pt}
\caption{Qualitative comparison on PKU unannotated dataset.}
\label{fig:pku_un_un}
\end{center}
\end{figure*}

\begin{figure*}[ht]
\centering
\includegraphics[width=0.8\linewidth]{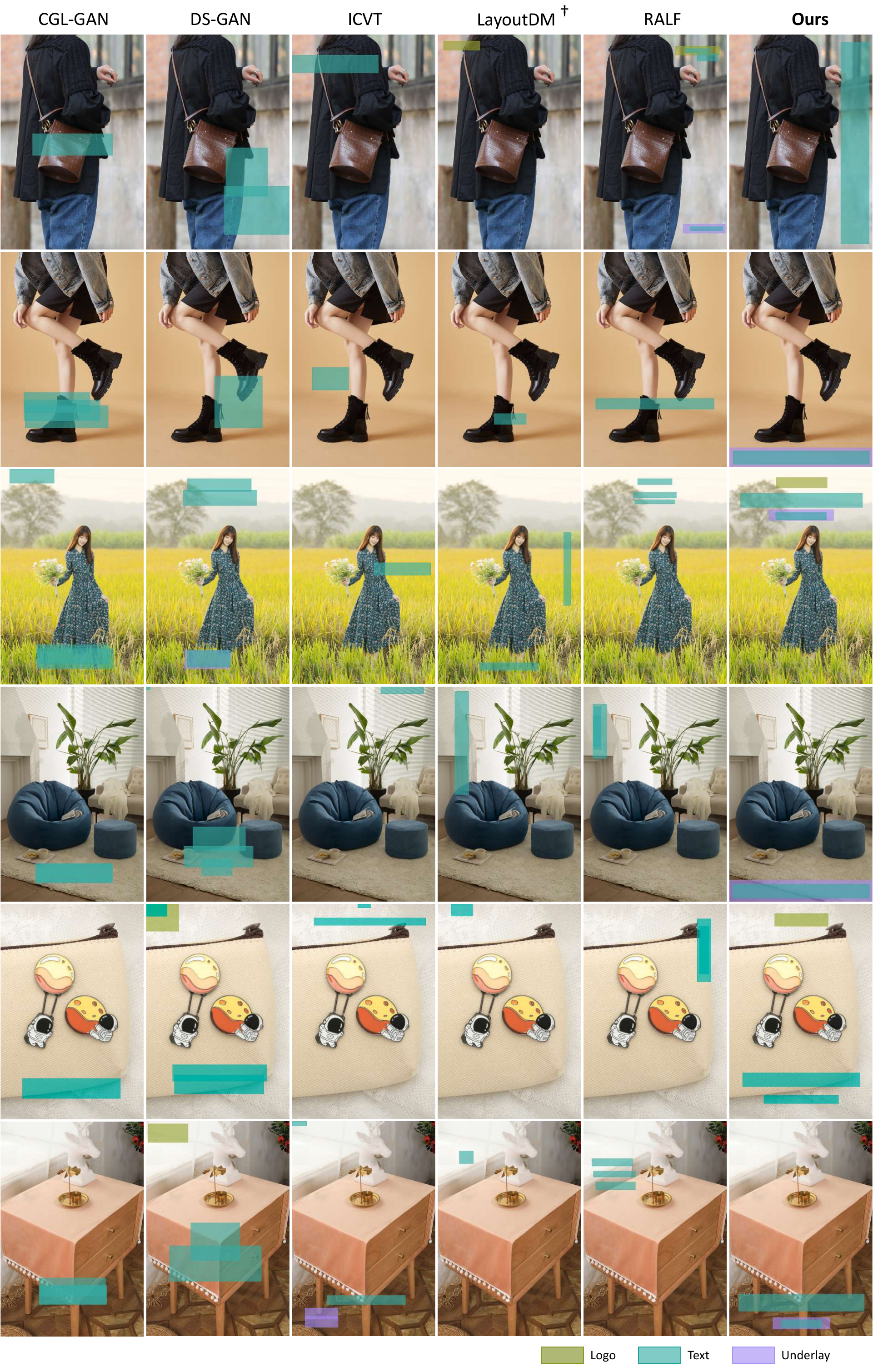}
%\vspace{-2pt}
\caption{Qualitative comparison on CGL unannotated dataset.}
\label{fig:cgl_un_un}
\end{figure*}

\begin{figure*}[ht]
\begin{center}
\centerline{\includegraphics[width=0.8\linewidth]{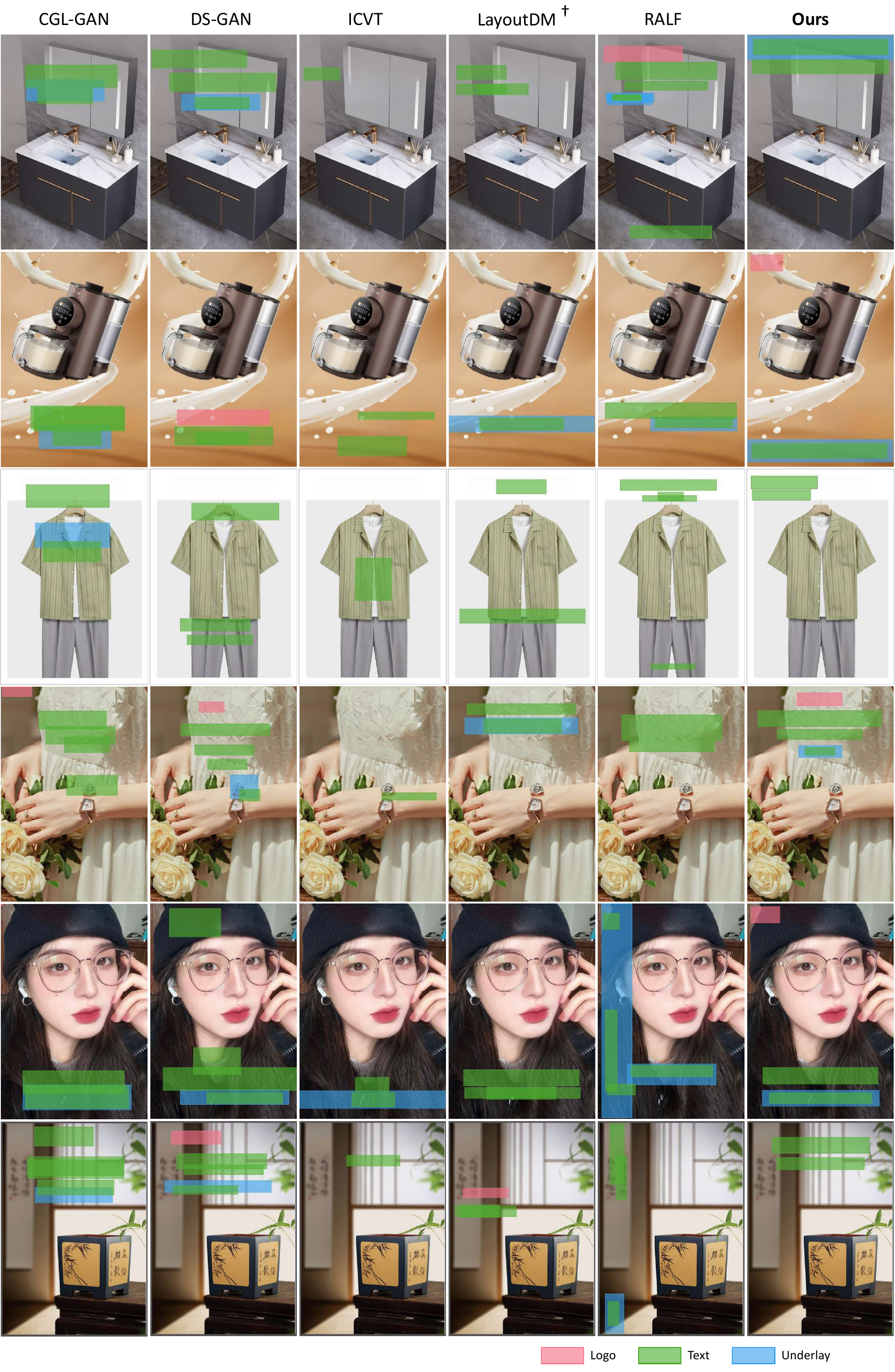}}
\vspace{-2pt}
\caption{Qualitative comparison on PKU annotated dataset.}
\label{fig:pku_un_an}
\end{center}
\end{figure*}

\begin{figure*}[ht]
\begin{center}
\centerline{\includegraphics[width=0.8\linewidth]{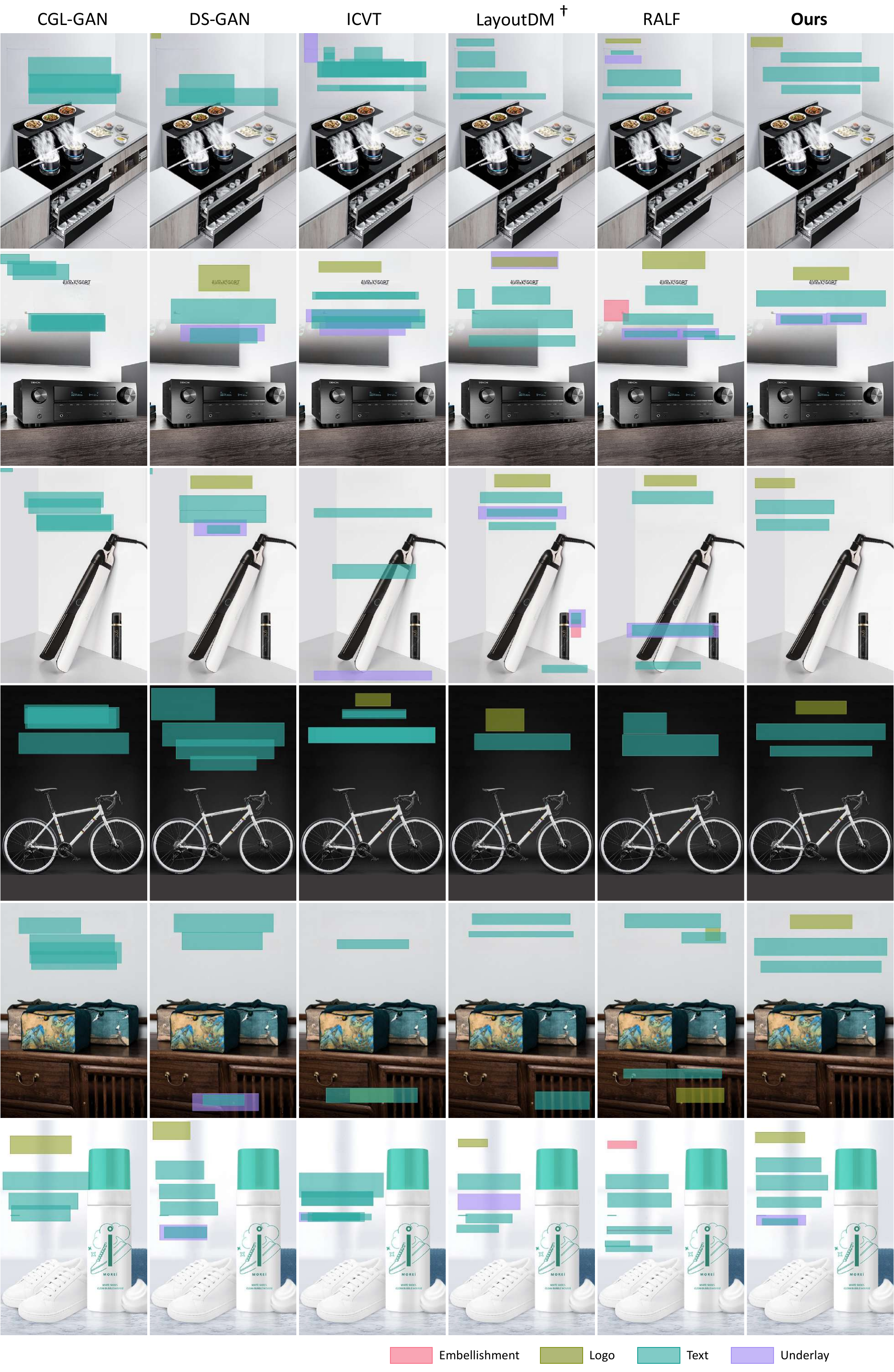}}
\vspace{-2pt}
\caption{Qualitative comparison on CGL annotated dataset.}
\label{fig:cgl_un_an}
\end{center}
\end{figure*}

We present additional qualitative comparisons in Fig.~\ref{fig:pku_un_un} and~\ref{fig:pku_un_an} for the PKU dataset without constraints, and in Fig.~\ref{fig:cgl_un_un} and~\ref{fig:cgl_un_an} for the CGL dataset. 

The results intuitively demonstrate that other baselines struggle to consistently generate reasonable layouts when confronted with diverse background images, exhibiting issues such as overlapping, misalignment, small-sized, and blocking.

\begin{figure*}[ht]
\begin{center}
\centerline{\includegraphics[width=1\linewidth]{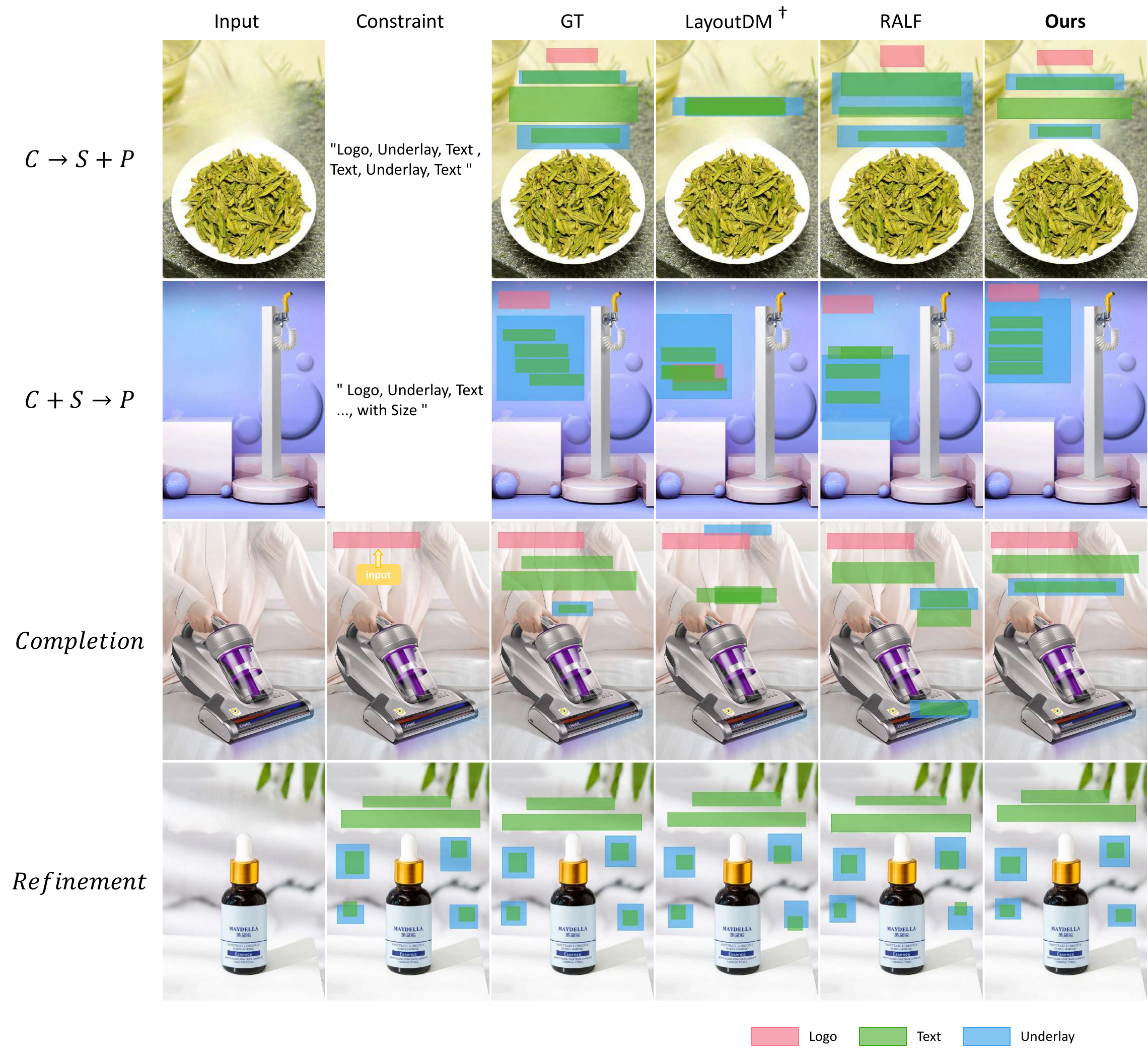}}
\vspace{-2pt}
\caption{Constrained qualitative comparison on PKU dataset.}
\label{fig:pku_con_an}
\end{center}
\end{figure*}

\begin{figure*}[ht]
\begin{center}
\centerline{\includegraphics[width=1\linewidth]{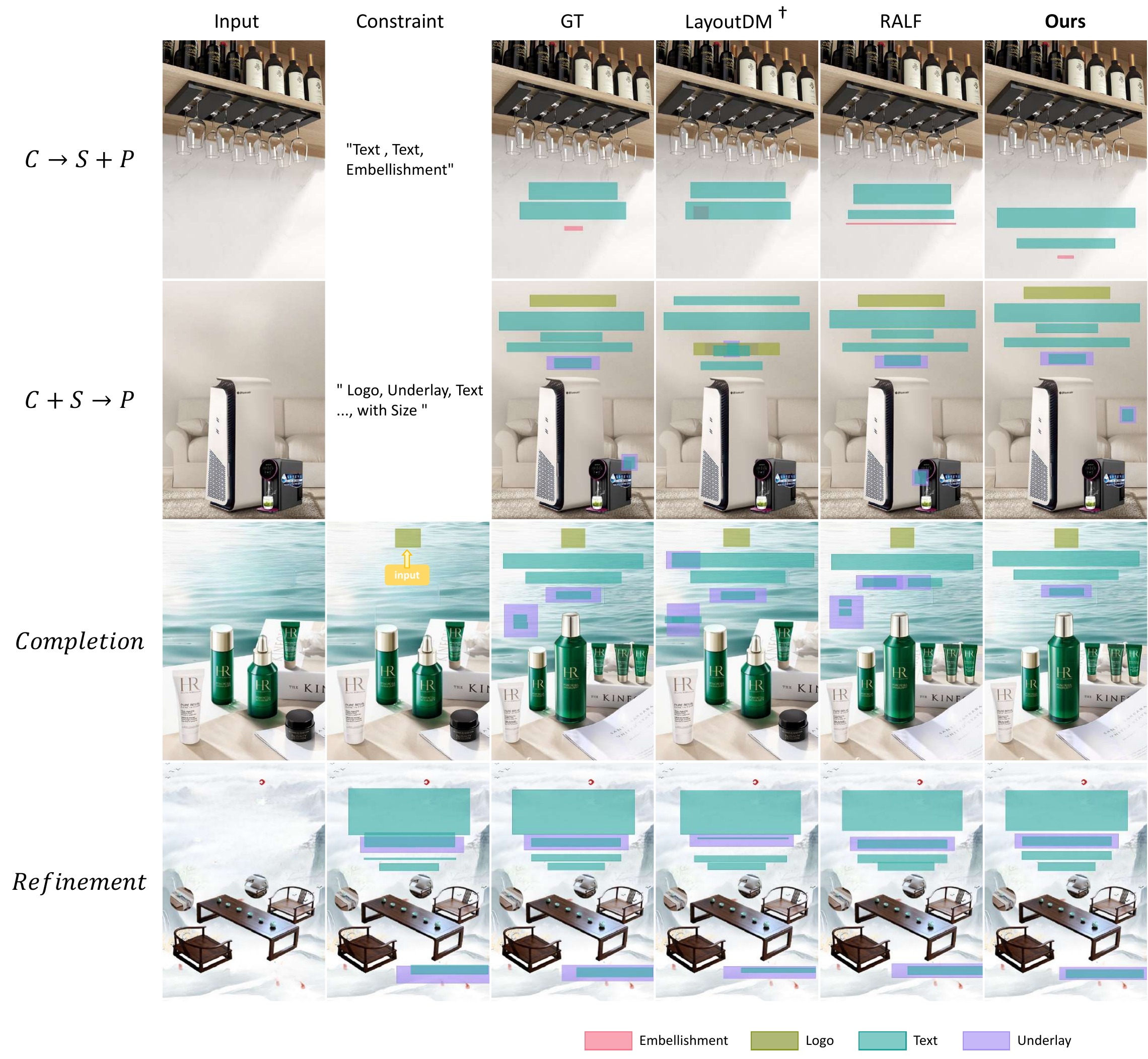}}
\vspace{-2pt}
\caption{Constrained qualitative comparison on CGL dataset.}
\label{fig:cgl_con_an}
\end{center}
\end{figure*}

Furthermore, in Fig.~\ref{fig:pku_con_an} and~\ref{fig:cgl_con_an}, we also present more results of constrained generation. The results indicate that even in constrained generation tasks, our method can generate high-quality layouts while strictly adhering to the given constraints without the need for additional constraint encoders.

\end{document}